\documentclass[lettersize,journal]{IEEEtran}
\usepackage{amsmath,amsfonts}

\usepackage{algorithm}
\usepackage{array}
\usepackage[caption=false,font=normalsize,labelfont=sf,textfont=sf]{subfig}
\usepackage{textcomp}
\usepackage{stfloats}
\usepackage{url}
\usepackage{verbatim}
\usepackage{graphicx}
\usepackage{cite}
\usepackage{CJK}
\usepackage{amssymb}
\usepackage{epstopdf}
\usepackage{algpseudocode}
\usepackage{multirow} 

\newcommand{\figref}[1]{Fig.~\ref{#1}}

\def\BibTeX{{\rm B\kern-.05em{\sc i\kern-.025em b}\kern-.08em
		T\kern-.1667em\lower.7ex\hbox{E}\kern-.125emX}}
\usepackage{balance}
\begin{document}

\title{FER-YOLO-Mamba: Facial Expression Detection and Classification Based on Selective State Space}

\author{Hui Ma, Sen Lei, Turgay Celik, \IEEEmembership{Member,~IEEE} and Heng-Chao Li, \IEEEmembership{Senior Member,~IEEE} 
	
\IEEEcompsocitemizethanks{
	\IEEEcompsocthanksitem Hui Ma, Sen Lei and Heng-Chao Li are with the School of Information Science and Technology, Southwest Jiaotong University, Chengdu 610031, China.\protect
	~E-mail: mahuiswjtu@163.com, senlei@swjtu.edu.cn, lihengchao\_78@163.com}

\IEEEcompsocitemizethanks{
\IEEEcompsocthanksitem Turgay Celik is with the School of Electrical and Information Engineering, University of the Witwatersrand, Johannesburg 2000, South Africa.\protect
~E-mail: celikturgay@gmail.com }

\thanks{Manuscript received May 6, 2024.}} 

\markboth{Journal of \LaTeX\ Class Files,~Vol.~X, No.~X, Apr.~2024}%
{Shell \MakeLowercase{\textit{et al.}}: A Sample Article Using IEEEtran.cls for IEEE Journals}

\IEEEpubid{0000--0000/00\$00.00~\copyright~2024 IEEE}

\maketitle

\begin{abstract}
Facial Expression Recognition (FER) plays a pivotal role in understanding human emotional cues. However, traditional FER methods based on visual information have some limitations, such as preprocessing, feature extraction, and multi-stage classification procedures. These not only increase computational complexity but also require a significant amount of computing resources. Considering Convolutional Neural Network (CNN)-based FER schemes frequently prove inadequate in identifying the deep, long-distance dependencies embedded within facial expression images, and the Transformer's inherent quadratic computational complexity, this paper presents the FER-YOLO-Mamba model, which integrates the principles of Mamba and YOLO technologies to facilitate efficient coordination in facial expression image recognition and localization. Within the FER-YOLO-Mamba model, we further devise a FER-YOLO-VSS dual-branch module, which combines the inherent strengths of convolutional layers in local feature extraction with the exceptional capability of State Space Models (SSMs) in revealing long-distance dependencies. To the best of our knowledge, this is the first Vision Mamba model designed for facial expression detection and classification. To evaluate the performance of the proposed FER-YOLO-Mamba model, we conducted experiments on two benchmark datasets, RAF-DB and SFEW. The experimental results indicate that the FER-YOLO-Mamba model achieved better results compared to other models. The code is available from https://github.com/SwjtuMa/FER-YOLO-Mamba.

\end{abstract}


\section{Introduction}

\IEEEPARstart{F}{acial} expression recognition (FER), as a fundamental component of emotion recognition, effectively captures and analyzes subtle facial changes to reveal an individual's emotional state. With the advancement of artificial intelligence (AI) and computer vision (CV), it has become a cornerstone in the field of affective computing, providing robust support for applications such as human-computer interaction and emotion analysis \cite{Li2022}. Accurate recognition of facial expressions not only allows for deeper insights into the complex connotations of human emotions but also establishes a strong foundation for developing intelligent and empathetic interaction systems. Currently, FER has been widely applied in affective computing, human-computer interaction, assistive healthcare, intelligent monitoring and security, the entertainment industry, remote education, and emotional state analysis, garnering considerable interest from numerous researchers~\cite{adyapady2023comprehensive}.

Traditional visual-based FER typically relies on visual information, such as facial images or videos, to analyze and recognize individuals' facial expressions and determine their emotional states. This technology draws from CV and pattern recognition, involving multiple steps such as preprocessing, feature extraction, and classification of facial images. For FER tasks, facial image preprocessing is essential for subsequent feature extraction and recognition, including face detection, alignment, and normalization operations. After preprocessing, specific algorithms or models are used to extract facial expression features, including shape, texture, and motion information from areas such as the eyes, mouth, and eyebrows. Based on the extracted features, a classifier or recognition algorithm is employed to classify facial expressions, including identifying smiles, anger, surprise, and other emotional states \cite{tian2001recognizing,subudhiray2023k}. Although vision-based FER technology has achieved a series of significant results, it often relies on manually designed feature extractors, which may to some extent limit its ability to accurately capture and classify complex and varied facial expressions. Due to the diversity and dynamics of facial expressions, traditional manual feature extraction methods may not comprehensively capture subtle facial changes, leading to a decrease in classification accuracy. Additionally, factors such as lighting conditions, head pose, and occlusions may also have a negative impact on recognition performance~\cite{Belmonte}.

\IEEEpubidadjcol

Moreover, deep learning-based object detection is capable of detecting objects while acquiring deep-level features, thus leading to precise classification. Consequently, there are extensive research prospects for applying this technology to the detection and classification of facial expressions. Current research on deep learning-based FER primarily focuses on the optimization of CNN and Transformer models, since CNN often struggles to capture long-distance dependency relationships and fine-grained facial expression features, and the Transformer model is constrained by quadratic computational complexity~\cite{Vaswani}. In this context, due to the excellent performance in modeling long-distance interactions while maintaining linear computational complexity, the Mamba model in state space models (SSMs) \cite{gu2023mamba} has led researchers to turn their attention to SSMs to address these limitations.

Aiming to overcome the limitations of existing technologies, this paper proposes a YOLO-Mamba model for FER tasks, named FER-YOLO-Mamba, which combines the advantages of YOLO and Mamba to achieve efficient detection and classification of facial expression images. The main contributions of this paper can be summarized as follows:
\begin{itemize}
	\item We innovatively develop a FER-YOLO-Mamba model, which establishes a visual backbone network grounded on the SSM. This represents a pioneering effort in integrating SSM-driven architectures into the realm of facial expression detection and classification, initiating an exploration of this model in this field.
	\item We further design a dual-branch structure that not only integrates the original local detailed information with the global contextual information provided by OSS but also incorporates an attention mechanism with a multi-layer perceptron. This attention mechanism integrates global average pooling, multi-layer perceptrons (MLPs), and element-wise multiplication techniques to implement a spatial attention mechanism for input feature maps. By selectively amplifying critical information regions while dampening the influence of irrelevant or secondary areas, this module significantly enhances the discriminative power and precision of the model in FER tasks.	
	\item To verify the effectiveness of the proposed FER-YOLO-Mamba model, we conducted experiments on two manually annotated facial expression datasets, RAF-DB and SFEW. The experimental results show that compared to other methods, the FER-YOLO-Mamba model achieved better results.
\end{itemize}
The remainder of this paper is organized as follows: Section \uppercase\expandafter{\romannumeral2} provides an overview of the related work. Section \uppercase\expandafter{\romannumeral3} gives the principles of SSM before delving into the design of our proposed FER-YOLO-Mamba. In Section \uppercase\expandafter{\romannumeral4}, we disclose the datasets employed for experimentation and provide the performance analysis. Lastly, Section \uppercase\expandafter{\romannumeral5} concludes the paper.

\section{Related work}
\label{sec:Related Work}
\subsection{Facial Expression Recognition}

FER plays a crucial role in the field of human-computer interaction, especially in applications like intelligent robots and virtual assistants. By accurately identifying the user's facial expressions, the system can better understand the user's emotions and intentions, thereby providing a more personalized service experience. Additionally, FER has demonstrated its unique value in the field of mental health. When assisting in the diagnosis and treatment of conditions such as depression and autism, doctors can more accurately assess the patient's emotional state by analyzing their facial expressions, leading to the development of more effective treatment plans. The application of this technology not only enhances the accuracy and efficiency of mental health services but also provides patients with a more precise diagnosis and treatment experience.

To achieve automatic classification of facial expressions, traditional visual information-based FER methods primarily focus on extracting and analyzing facial features through image processing techniques and pattern recognition algorithms. These methods typically involve face detection, feature extraction, and expression classification. 

During the feature extraction phase, traditional FER methods often rely on hand-crafted feature extractors, including geometric, texture, and motion feature extraction. In geometric feature-based approaches, features are obtained by analyzing geometric information such as the position, distance, and angle of facial landmark points. Tian et al. \cite{tian2001recognizing} proposed a geometric feature-based FER method that identifies and analyzes facial action units to achieve expression classification. Instead, texture feature-based methods utilize the changes in facial skin texture to recognize expressions, typically calculated through grayscale co-occurrence matrices or local binary patterns. Shan et al. \cite{shan2009facial} used local binary patterns as texture features for FER. Additionally, motion feature methods are used to capture facial muscle movements and changes to identify different expressions. Bartlett et al. \cite{bartlett2006automatic} proposed a FER method that combines geometric and dynamic features, achieving good performance in spontaneous expression recognition.

In classifier design, traditional methods often utilize machine learning algorithms such as Support Vector Machines (SVM), K-nearest neighbors (KNN), {\it etc}. Alhussan et al. \cite{alhussan2023facial} presented an effective method for FER based on optimized SVM, emphasizing the significance of model optimization and feature extraction in enhancing recognition performance. Subudhiray et al. \cite{subudhiray2023k} discussed facial emotion recognition technology based on the K-Nearest Neighbors (KNN) algorithm, underlining the importance of using effective features. 

To overcome the limitations of hand-crafted feature extractors in FER, these traditional methods often struggle to comprehensively capture critical information closely related to expressions and lack robustness to changes in lighting, facial poses, and occlusions. Therefore, an increasing number of researchers are turning to deep learning models, especially CNNs, for facial expression recognition tasks. As a result, Wang et al.~\cite{wang2021information} introduced a CNN-based FER method and focused on the concept of information reuse attention. They designed a network structure to promote information sharing and reuse between different convolutional layers in complex environments. Sarvakar et al. \cite{sarvakar2023facial} built a CNN-based FER model, which is trained and tested using multiple facial expression datasets. Additionally, Patro et al.~\cite{Patro} developed a FER system based on a customized DCNN. Through deep learning methods, the system can automatically learn and extract features related to different emotions, such as happiness, sadness, anger, etc., from facial images.

In addition, as another ongoing area of FER research, multimodal information fusion-based FER methods incorporate not only visual information but also integrate multimodal sources such as audio, text, and others to further enhance the accuracy and reliability of FER. Zadeh et al. \cite{zadeh2017tensor} studied the utilization of different data sources for emotion analysis, where the Tensor Fusion Network was proposed to integrate and analyze data from different modalities. Similarly, Pan et al. \cite{pan2023multimodal} proposed a multimodal emotion recognition method based on facial expressions, speech, and electroencephalogram (EEG). The extracted emotion features included not only traditional features of facial expressions and speech but also characteristics in EEG signals. Also, in \cite{zhang2023deep}, Zhang et al. provided a systematic review of deep learning-based multimodal emotion recognition techniques, primarily discussing the latest developments and prospects in emotion recognition.

\subsection{Object Detection Methods Based on the YOLO Series}

Recently, with the advancement of deep learning technology, significant progress has been achieved in object detection algorithms. Among these algorithms, the YOLO (You Only Look Once) series algorithms have gained widespread attention due to their efficiency and real-time performance.

The first version of the YOLO algorithm \cite{7780460} introduced the concept of transforming the object detection task into a regression problem. Subsequently, YOLOv2 \cite{8100173} brought several improvements over the original version, including the introduction of batch normalization to enhance the model's convergence speed and stability, as well as a high-resolution classifier to improve its capability to capture fine-grained features. Furthermore, YOLOv3 \cite{Redmon2018YOLOv3AI} improved the network structure by utilizing a deeper Darknet-53 architecture and introducing residual connections to prevent gradient disappearance and model degradation. YOLOv3 also adopted multi-scale prediction to effectively capture objects of different sizes by detecting them on feature maps of varying scales. 

Subsequently, aiming for higher accuracy while preserving the efficiency of the YOLO series, YOLOv4 \cite{Bochkovskiy2020YOLOv4OS} emerged, employing a more complex network structure, CSPDarknet53, and introducing techniques such as Cross-Stage Partial connections (CSP) and Self-Adversarial Training (SAT). YOLOv5 improved the model's flexibility and usability by adopting more efficient calculation methods and hardware acceleration techniques, allowing for high accuracy and fast detection speed. By dividing the grid and predicting the position and class of each object in every grid, YOLOv7 \cite{Wang2022YOLOv7TB} achieved rapid and accurate object detection. Compared to previous versions, YOLOv7 has improved detection accuracy and can meet the requirements of more application scenarios. As the latest model in the YOLO series, YOLOv8 was released by Ultralytics and built upon the historical versions of the YOLO series. With the introduction of new features, YOLOv8 utilized a deeper, more complex network structure, as well as more efficient loss functions, resulting in higher detection accuracy and faster detection speed. Furthermore, the YOLOX \cite{Ge2021YOLOXEY} object detection algorithm, developed by Megvii, represents an advancement built upon YOLOv3-SPP. It has transformed the original anchor-based approach into an anchor-free form and incorporates other advanced detection technologies, such as decoupled head and label assignment SimOTA, resulting in outstanding performance.

\subsection{State Space Model on Visual Recognition}

The State Space Model (SSM) has recently gained prominence in deep learning as a pivotal method for state space transformation \cite{Smith2022SimplifiedSS}. Drawing inspiration from the SSM in continuous control systems and integrating the cutting-edge HiPPO initialization method \cite{Gu2020}, the LSSL model \cite{Gu2021Gu} has effectively demonstrated the extensive potential of the SSM in addressing long-term dependencies in sequences. However, the LSSL model faces constraints due to the computational complexity of state representation and substantial storage requirements. To address this, the S4 model \cite{gu2022efficiently} was introduced to enhance performance through parameter diagonal structuring and normalization. Subsequently, a series of SSMs with diverse structures (e.g., complex diagonal structure \cite{gupta2022diagonal}, selective mechanisms, and others \cite{gu2023mamba}) have emerged, showcasing significant advantages in their respective application scenarios.

In visual processing, Liu et al. \cite{liu2024vmamba} drew inspiration from the SSM, proposing the Visual State Space Model (VMamba). This model not only inherits the advantages of SSM in the global receptive field but also achieves linear computational complexity, significantly improving the efficiency of image processing. Subsequently, by introducing the Res-VMamba model, Chen et al. \cite{chen2024res} further enhanced the VMamba model and optimized it specifically for fine food image classification tasks. In remote sensing image classification, Chen et al. \cite{chen2024rsmamba} put forth the RSMamba model, harnessing an efficient, hardware-aware Mamba implementation to effectively integrate the advantages of global receptive field and linear complexity modeling. 

While in medical image processing, Yue et al. \cite{yue2024medmamba} introduced the MedMamba model, the first specific Mamba model designed for medical image classification. Additionally, Ma et al. \cite{ma2024u} proposed the U-Mamba model, which effectively enhances the performance of biomedical image segmentation by combining the advantages of the U-Net architecture and the Mamba model. The VM-UNet model, proposed by Ruan et al. \cite{ruan2024vm}, combines Vision Mamba with U-Net for medical image segmentation tasks, bolstering segmentation accuracy and robustness through integrated multi-scale feature information. Liu et al. \cite{liu2024swin} presented the Swin-UMamba model, which integrates the Swin Transformer into Mamba for pre-training, further contributing to the model's accuracy in biomedical image segmentation tasks. Furthermore, Yang et al. \cite{yang2024vivim} introduced the Vivim model, offering a novel approach for medical video object segmentation. Gong et al. \cite{gong2024nnmamba} showcased the remarkable performance of the nnMamba model, which demonstrates excellent performance in handling complex 3D image data by combining deep learning with the benefits of SSM. Finally, Guo et al. \cite{guo2024mambamorph} proposed the MambaMorph model, providing a new solution for deformable MR-CT registration tasks. 

\section{Methodology}
\label{sec:Proposed Model}
\subsection{State Space Models}

State Space Models (SSMs) have garnered increasing favor among researchers due to their unique capability to encapsulate dynamic systems. This type of model can effectively transform input sequences, represented as $x(t) \in \mathbb{R}^{L}$, into output variables, denoted as $y(t) \in \mathbb{R}^{L}$, through implicit latent states $h(t) \in \mathbb{R}^N$, showcasing robust adaptability in modeling complex time series. SSMs are deeply rooted in control theory, with its core structure being represented by a set of linear ordinary differential equations (ODEs) as follows:
\begin{equation}
\begin{array}{l}
h^{\prime}(t)=\mathbf{A} h(t)+\mathbf{B} x(t), \\
y(t)=\mathbf{C} h(t)+\mathbf{D} x(t),
\end{array}
\end{equation}
where $\textbf{A} \in \mathbb{C}^{N \times N}, \mathbf{B}, \mathbf{C} \in \mathbb{C}^N $ for a state size $N$, and the skip connection $ \mathbf{D} \in \mathbb{C}^1$.

In SSMs, the state transition matrix \textbf{A} plays a crucial role in governing the evolution path of the state vector $h(t)$, while the input matrix \textbf{B}, output matrix \textbf{C}, and feedforward matrix \textbf{D} respectively reveal the intrinsic connections between the input signal $x(t)$, the state $h(t)$, and the output response $y(t)$. In deep learning, there is often a preference for adopting a discrete-time framework, which requires the transformation of the continuous equations describing the dynamic characteristics of the system into a discrete form to meet computational requirements and ensure synchronization with the sampling frequency of data acquisition.

The discretization of SSMs essentially transforms the system's continuous-time system of ordinary differential equations into an equivalent discrete-time representation, which can be achieved by applying a zero-order hold strategy to the input signal, thereby constructing the discrete-time SSM as follows:
\begin{equation}
\begin{aligned}
h_k & =\bar{\mathbf{A}} h_{k-1}+\bar{\mathbf{B}} x_k, \\
y_k & =\bar{\mathbf{C}} h_k+\bar{\mathbf{D}} x_k,
\end{aligned}
\end{equation}         
where $\bar{\mathbf{A}} =e^{\mathbf{\Delta A}}, \bar{\mathbf{B}} =\left(e^{\mathbf{\Delta A}}-{\textbf{I}}\right) {\mathbf{A}}^{-1} \mathbf{B}, \bar{\mathbf{C}}={\mathbf{C}}$, $ \mathbf{B},\mathbf{C}\in\mathbb{R}^{D \times N}$, and $\mathbf{\Delta} \in \mathbb{R}^D $.

The Mamba algorithm~\cite{gu2023mamba}, with its unique selective scanning mechanism within the SSM framework, demonstrates significant advantages in facial expression detection and classification tasks. The core of this mechanism lies in its ability to dynamically adjust the system matrices $\mathbf{B}$ and $\mathbf{D}$ based on the current and historical context, a key feature that sets it apart from other methods.

In facial expression image analysis, diversity and complexity pose challenges to traditional methods. However, the Mamba algorithm, through its selective scanning mechanism, focuses on key areas of input data, effectively extracting features relevant to facial expressions. This precise focus allows the algorithm to more accurately capture subtle changes in expressions, thereby improving detection and classification accuracy.

More importantly, the Mamba algorithm enhances its ability to handle complex temporal dynamics by dynamically adjusting the system matrices $\mathbf{B}$ and $\mathbf{D}$. This is particularly crucial in facial expression detection and classification, as facial expressions involve not only subtle differences within a single frame but also dynamic changes between consecutive frames. The algorithm can respond in real-time to the changing characteristics of input data, accurately capturing this complex temporal dynamics, thus better understanding the continuity and dynamics of facial expressions and enhancing the accuracy of detection and classification.

In conclusion, the Mamba algorithm, with its unique selective scanning mechanism and dynamic adjustment capabilities, shows great potential in facial expression detection and classification tasks. Its advantages in capturing subtle expression changes and dynamic features make the algorithm have broad application prospects and significant research value in the field of facial expression image analysis.

\subsection{Overall architecture}
\begin{figure*}[!t]
	\centering
	\includegraphics[width=6.2in]{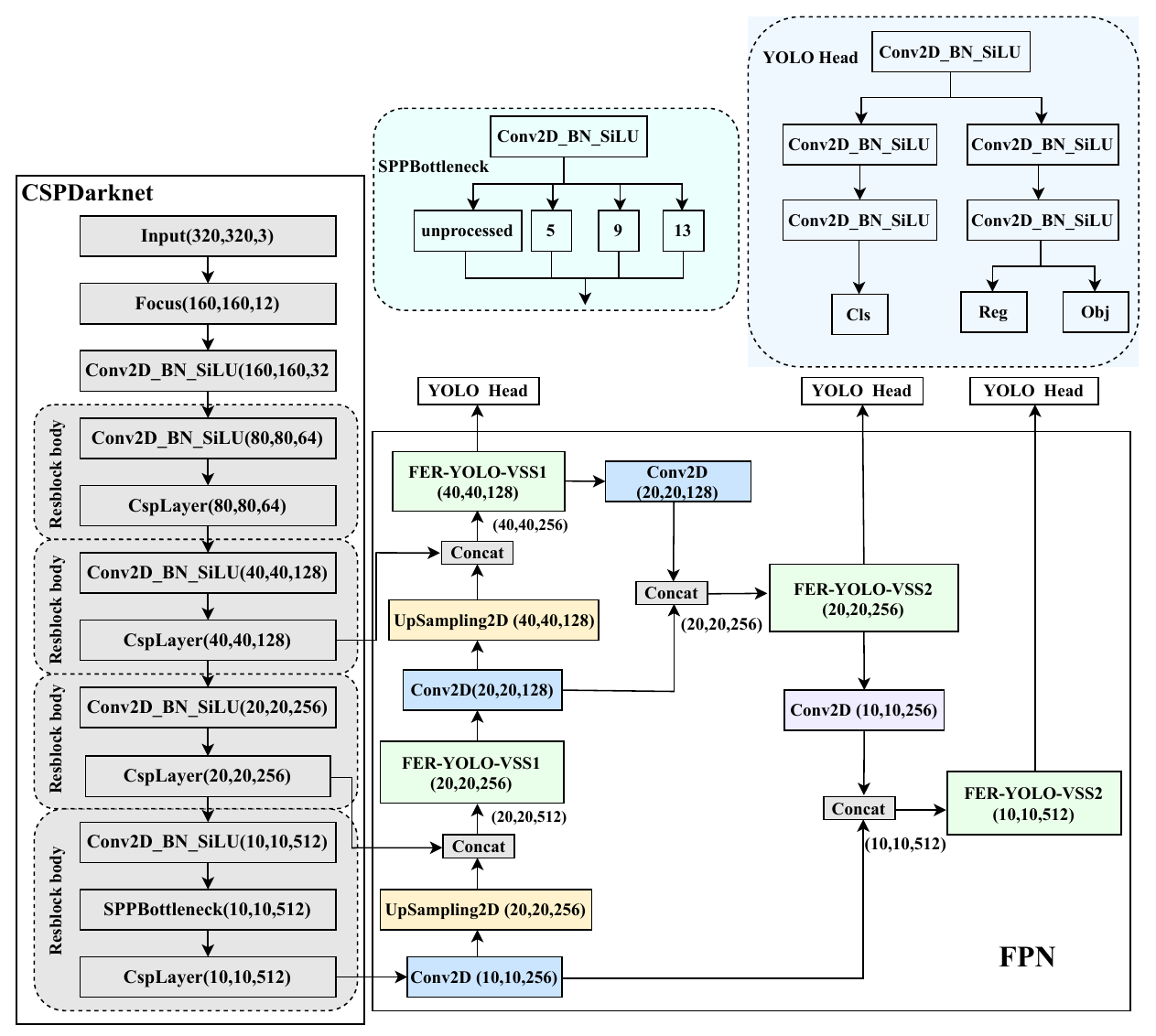}
	\caption{The overall architecture of the FER-YOLO-Mamba.}
	\label{fig01}
\end{figure*}

\figref{fig01} shows the architecture of the proposed FER-YOLO-Mamba network, which mainly consists of three core parts: CSPDarknet, FPN, and YOLO Head. Initially, CSPDarknet serves as the backbone feature extraction network, responsible for the initial feature extraction from the input image. After the processing by CSPDarknet, the input image is transformed into three feature maps of different scales, with dimensions of $10\times10\times512$, $20\times20\times256$ and $40\times40\times128$, containing hierarchical multi-level feature information from coarse to fine.

The FPN serves as an enhanced feature extraction network by integrating the multi-scale features outputted by CSPDarknet. The central concept of this module lies in effectively fusing cross-scale features to capture details and context information at different levels, thereby enhancing the overall feature representation. Specifically, FPN applies upsampling to upsample the low-level feature maps to match the dimensions of the high-level feature maps for cross-scale interaction, while also implementing downsampling operations to enrich the dimensions and depth of feature fusion.

As a pivotal component of the FER-YOLO-Mamba framework, the YOLO Head shoulders the dual responsibilities of classification and localization. After the collaborative processing by CSPDarknet and FPN, the network generates three reinforced multi-scale feature maps. These feature maps can be envisioned as grids comprising a large number of feature points, each harboring a feature vector associated with its channels. The core mechanism of the YOLO Head involves analyzing these feature points individually to ascertain their association with a target object. This process comprises two complementary and independent subtasks: class prediction to determine the target class linked to feature points, and bounding box regression to precisely estimate the target's position. Ultimately, the outputs of these two types of predictions are fused to comprehensively identify the targets in the image.

Compared to conventional object detection datasets, FER datasets have unique characteristics. Although they focus only on one feature, they are often disturbed by complex backgrounds. Traditional FER methods often use preprocessing techniques to weaken the background influence and simplify the recognition process. However, in the design of the FER-YOLO-Mamba model, we did not adopt such preprocessing steps. Instead, we directly used the original images with background as input. These input images have dimensions of $320\times320\times3$ and are rich in background information, which undoubtedly places higher demands on the model's ability to handle complex scenes and interference. At the same time, this also underscores the unique strategy employed by the FER-YOLO-Mamba model in addressing FER tasks with complex backgrounds, and its immense potential for practical applications.

\subsection{FER-YOLO-VSS module}

\begin{figure*}[!t]
	\centering
	\includegraphics[width=6in]{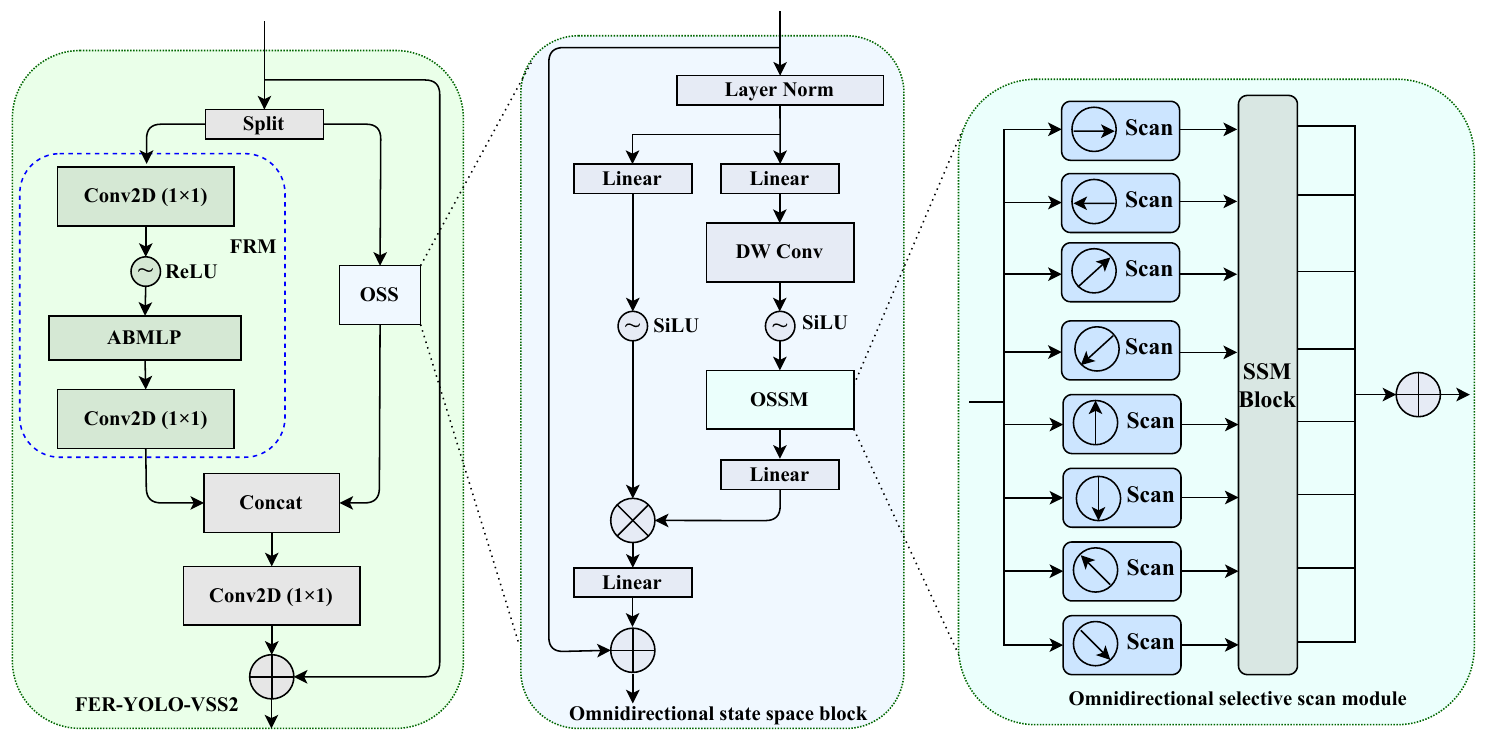}
	\caption{FER-YOLO-VSS2 module.}
	\label{fig02}
\end{figure*}

The FER-YOLO-VSS module is a dual-branch structure. Specifically, the input of this module is first processed through channel splitting, divided into two equally sized sub-inputs for independent feature extraction and processing in subsequent steps. This design aims to more effectively capture and extract key feature information in images through a parallel processing strategy. Subsequently, these two sub-inputs enter their respective specific processing branches, namely the Feature Refinement Module (FRM) branch and the Omnidirectional State Space (OSS) branch.

To enhance the model's ability to learn discriminative and context-aware feature representations, the FRM branch adopts a continuous channel dimension compression strategy. Additionally, this branch incorporates an attention mechanism with adaptive feature weight adjustment to adjust the importance of different features. After this series of processing, the FRM branch eventually restores the original number of channels, thereby ensuring the integrity and accuracy of information.

The OSS branch \cite{zhao2024rs} initiates with layer normalization applied to the input features as a preprocessing step, following which the normalized features are divided into two parallel sub-paths. In the first path, the features undergo a simplified transformation comprising linear transformation layers and activation functions. Meanwhile, the second path involves a relatively more complex process, as the features undergo three levels of progressive processing including linear layers, depthwise separable convolutions, and activation functions, before entering the Omnidirectional Selective Scan Module (OSSM) to deeply extract feature information.

The OSSM utilizes SSM technology to implement bidirectional selective scanning of facial expression images in the horizontal, vertical, diagonal, and reverse diagonal directions. This approach aims to enhance the global effective receptive field of the images in multiple directions and extract global spatial features from various perspectives. Specifically, selective scanning in eight different directions enables the capture of large-scale spatial features from multiple orientations. Following this, layer normalization is applied to standardize the features, and the output of this branch is deeply fused with the output of the first branch through element-wise multiplication. Subsequently, with the assistance of linear blending layers to integrate features from each branch and the incorporation of residual connection strategies, the final output response of the FER-YOLO-VSS2 module is synergistically constructed. Within the OSS branch, the SiLU activation function is chosen as the default activation unit. Finally, the output characteristics of the two branches are concatenated along the channel dimension, and deep feature fusion is performed through a 1x1 convolutional layer to enhance the deep-level interaction effects between feature maps.

Based on the difference in output channel numbers, the FER-YOLO-VSS module is divided into two variants: FER-YOLO-VSS1 and FER-YOLO-VSS2~(\figref{fig02}). FER-YOLO-VSS1 aims to reduce the number of channels and does not introduce a ``shortcut'' connection mechanism ($C \to \frac{C}{2}$). At the same time, FER-YOLO-VSS2, maintaining the consistency between input and output channel numbers, incorporates a ``shortcut'' connection to optimize the efficiency of information flow ($C \to C$). 

Overall, as the core module, FER-YOLO-VSS integrates not only the original local information but also the global contextual information provided by OSS. It also incorporates an attention mechanism with multi-layer perceptrons. This design strategy aims to achieve complementary fusion of local and global information, enhancing the model's ability to process key information through the attention mechanism, thereby improving overall performance.

\subsection{Attention Block with Multi-Layer Perceptron}

The Attention Block with Multi-Layer Perceptron (ABMLP) module integrates global average pooling, multi-layer perceptron (MLP), and element-wise multiplication techniques to implement a spatial attention mechanism for the input feature map. Its core function is to selectively highlight key information areas while attenuating the influence of irrelevant or minor areas, thereby enhancing the discriminative performance of the model in recognition tasks.

The pseudocode for ABMLP is illustrated in Algorithm \ref{alg_attention_block}. Initially, the input features map, $x \in \mathbb{R}^{b \times c \times h \times w}$ undergoes global average pooling to obtain a feature vector, $y$, with a shape of $(b,c)$. Subsequently, the feature vector, $y$, is fed into an MLP to generate attention-weight vectors through a series of non-linear transformations. This MLP consists of three linear layers, incorporating ReLU activation functions after two layers to introduce non-linear characteristics, and a Sigmoid activation function at the end to produce the attention weight vectors.

This weight vector is appropriately reshaped to match the original input feature map's dimensions, resulting in a shape of $(b,c,1,1)$ for further operations. Finally, the reshaped attention weight vector is element-wise multiplied with the original input feature map, $x$, to generate a self-attention-enhanced feature map, which is the output result of the ABMLP module.
\begin{algorithm}[!htbp]
	\caption{ABMLP Pseudo-Code}
	\label{alg_attention_block}
	\begin{algorithmic}[1]
		\Require Input: $x \in \mathbb{R}^{b \times c \times h \times w}$
		\Ensure Output: Attention-augmented feature map
		
		\State $y \gets \text{GlobalAveragePooling2D}(x)$ \Comment{Step 1}
		\State $y \gets \text{Flatten}(y)$ \Comment{Step 2}
		\State $y \gets \text{MLP}(y) \to \text{Sigmoid}$ \Comment{Step 3}
		\State $y \gets \text{Reshape}(y, \text{shape}=(b, c, 1, 1))$ \Comment{Step 4}
		\State $\text{Output} \gets x \times y$ \Comment{Step 5}
		
	\end{algorithmic}
\end{algorithm}

\section{Experimental Results and Analysis}
\label{sec:Experimental Results and Analysis}
\subsection {Datasets and Implementation Details}

1) \emph{Facial expression datasets:} in this paper, we conducted experiments based on two facial expression datasets, RAF-DB~\cite{Li2017} and SFEW~\cite{SFEW}.

The RAF-DB dataset is a large-scale FER dataset that consolidates images from diverse real-life scenarios, such as social media visuals and movie frames, vividly illustrating the complexity and diversity of expression recognition in natural settings. The dataset covers seven basic expressions as well as 21 compound expressions, while in this paper, experiments are limited to the seven basic expressions. The dataset comprises 12,271 training images and 3,068 testing images.

The SFEW dataset serves as a benchmark specifically designed for research on FER in complex real-world environments. A notable feature of this dataset is the in-the-wild nature of expressions, occurring in natural and uncontrolled scenes. Derived from the AFEW video database, the dataset is meticulously annotated with key facial expression frames, comprising 1,251 images that depict various lighting conditions, background complexities, head poses, and facial occlusions, accurately simulating the complex scenarios encountered in real-world expression recognition tasks. The dataset also includes seven basic facial expressions.

For the end-to-end experiment, we employed manual labeling for both datasets to ensure the accuracy and consistency of the labels. Note that no form of preprocessing, such as facial alignment, was carried out during the experimental process. This deliberate approach was employed to assess the model's performance under the condition of non-standardized input data.

2) \emph{Implementation Details:} the experiments were performed on a server platform with specific hardware configurations, utilizing the PyTorch framework for algorithm development and model training. The hardware specifications include an AMD Ryzen Threadripper 3960X 24-Core Processor, paired with 125 GB of memory and a GeForce RTX 2080 Ti graphics card, ensuring an efficient computing environment. In terms of training strategy, all models uniformly utilized the Adam optimization algorithm, with a batch size of 16 to balance computational efficiency and memory usage.

To address the variances across datasets, we tailored the training durations accordingly, setting 300 epochs for the RAF-DB dataset and extending this to 500 epochs for the more intricate SFEW dataset, thereby accommodating disparities in both data volume and intricacy. An initial learning rate of 0.001 was adopted, accompanied by an exponential decay strategy, implemented every 64 epochs throughout each training period. This strategy involved reducing the learning rate by a factor of 0.9 at each interval, strategically facilitating a gradual convergence towards the most favorable solution.

It is noted that the backbone feature extraction network loaded pre-trained weights from the COCO dataset before commencing training.

\subsection {Evaluation Metrics}
For the proposed FER-YOLO-Mamba model, we evaluated its performance based on a series of key performance metrics as follows:
\begin{itemize}
	\item {\it Precision}: this metric aims to measure the proportion of actual positive samples in the model's outputs that are predicted as positive, and it can be formulated as follows:
	\begin{equation} 
	Precision=\frac{TP}{TP+ FP}, 
	\end{equation} 
	where true positives ({\it TP}) represents the number of samples correctly identified as positive by the model, while false positives ({\it FP}) indicates the number of actual negative samples incorrectly classified as positive.
	
	\item {\it Avg\_Precision}: the mean value of $Precision$ across all classes. 
	\begin{equation} 
	Avg\_Precision=\frac{1}{N} \sum_{i=1}^{N} Precision_{i},
	\end{equation}
	where $Precision_{i}$ represents the $Precision$ of the $i$-th class, and $N$ is the total number of classes.
		
	\item {\it Recall}: {\it Recall} is used to measure the accuracy of the model in identifying all true positive samples, {\it i.e.}, the proportion of actual targets that are successfully detected by the model, and it can be formulated as follows:
	\begin{equation} 
		Recall = \frac{TP}{TP+FN}, 
	\end{equation} 
	where false negatives ({\it FN}) refers to the number of samples that are actually positive samples but are mistakenly classified as negative by the model.

	\item {\it Avg\_Recall}: the mean value of $Recall$ across all classes. 
	\begin{equation} 
	Avg\_Recall=\frac{1}{N} \sum_{i=1}^{N} Recall_{i},
	\end{equation}
	where $Recall_{i}$ represents the $Recall$ of the $i$-th class.
	
	\item {\it F1 score}: as the harmonic mean of {\it Precision} and {\it Recall}, the {\it F1 score} provides a single evaluation metric that balances both. A higher {\it F1 score} indicates that the model can effectively control the increase in {\it FP} rate while maintaining a high {\it Recall}.
	 {\it F1 score} is calculated as follows:
	\begin{equation} 
	F1\text{ \it score}=2 \cdot \frac{Precision \cdot Recall}{Precision+ Recall}. 
	\end{equation}
	
	\item {\it Avg\_F1}: the mean value of {\it F1 score} across all classes. 
	\begin{equation} 
	Avg\_F1=\frac{1}{N} \sum_{i=1}^{N} F1 \_ {score }_{i},
	\end{equation}
	where $ F1\_{score }_{i}$ represents the {\it F1 score} of the $i$-th class.
	
	\item {\it Average Precision (AP)}: for any given class, the {\it AP} aims to reflect the average performance of {\it Precision} at various {\it Recall} levels. The {\it AP} directly indicates the model's ability to maintain high $Precision$ at different $Recall$ levels. 
	
	\item{\it mAP}: {\it mAP} as the arithmetic average of all class {\it AP} values, is used to evaluate the overall performance of the model in multi-class detection tasks. {\it mAP} is calculated as follows:
	\begin{equation} 
	mAP=\frac{1}{N} \sum_{i=1}^N AP_i, 
	\end{equation} 
	where $AP_i$ represents the {\it AP} score for the $i$-th class.
\end{itemize}

\subsection{Comparisons with state-of-the-art methods}

\begin{table*}
	\centering
	\caption{Performance comparison of different methods on the RAF-DB dataset.}
	\label{table_RAF_result}
\begin{tabular}{|c|c|c|c|c|c|c|c|c|c|}
	\hline \multirow{2}{*}{ Method } & \multirow{2}{*}{ Year } & \multicolumn{7}{c|}{ $AP$(\%)} & \multirow{2}{*}{ $mAP$(\%) }  \\
	\cline { 3 - 9 } & & Anger & Disgust & Fear & Happy & Neutral & Sad & Surprise &  \\
	\hline SSD~\cite{Wei} & 2015 & $\underline{81.23}$ & $\underline{62.91}$ & 57.01 & 95.72 & 80.34 & 78.32 & $\underline{89.71}$ & 77.89   \\
	\cline { 1 - 10} RetinaNet~\cite{Lin2017} & 2017 & $\mathbf{82.07}$ & 53.74 & 53.56 & 94.63 & 80.13 & 77.50 & 87.80 & 75.63   \\
	\cline { 1 - 10} YOLOv3~\cite{Redmon2018YOLOv3AI} & 2018 & 58.01 & 28.56 & 37.02 & 88.09 & 67.89 & 63.78 & 72.59 & 59.42  \\
	\cline { 1 - 10} CenterNet~\cite{Xingyi} & 2019 & 53.26 & 17.41 & 30.62 & 91.32 & 75.36 & 66.83 & 84.63 & 59.92   \\
	\cline { 1 - 10} EfficientNet~\cite{Mingxing} & 2019 & 68.72 & 52.25 & 45.47 & 93.75 & 78.67 & 76.96 & 84.31 & 71.45    \\
	\cline { 1 - 10} YOLOv4~\cite{Bochkovskiy2020YOLOv4OS} & 2020 & 39.25 & 0.00 & 10.36 & 87.61 & 52.77 & 45.72 & 59.91 & 42.23    \\
	\cline { 1 - 10} YOLOv5 & 2020 & 45.75 & 8.86 & 0.00 & 91.77 & 64.73 & 65.60 & 74.36 & 50.15  \\
	\cline { 1 - 10} YOLOvX~\cite{Ge2021YOLOXEY} & 2021 & 78.38 & 62.40 & $\underline{57.85}$ & $\underline{96.82}$ & $\underline{80.45}$ & $\underline{83.35}$ & 89.56 & $\underline{78.40}$ \\
	\cline { 1 - 10} YOLOv7~\cite{Wang2022YOLOv7TB} & 2022 & 62.20 & 55.80 & 44.72 & 92.01 & 73.20 & 74.72 & 74.57 & 68.17 \\
	\cline { 1 - 10} YOLOv8 & 2023 & 74.50 & 50.40 & 50.85 & 93.33 & 76.39 & 76.30 & 82.89 & 72.09   \\
	\cline { 1 - 10} FER-YOLO-Mamba & 2024 & 79.55 & $\mathbf{64.32}$ & $\mathbf{62.00}$ & $\mathbf{97.43}$ & $\mathbf{83.23}$ & $\mathbf{84.22}$ & $\mathbf{91.44}$ & $\mathbf{80.31}$  \\
		\hline
\end{tabular}
\end{table*}
	
\begin{table*}
	\centering
	\caption{Performance comparison of different methods on the SFEW dataset.}
	\label{table_sfew_result}
	\begin{tabular}{|c|c|c|c|c|c|c|c|c|c|}
		\hline \multirow{2}{*}{ Method } & \multirow{2}{*}{ Year } & \multicolumn{7}{c|}{ $AP$(\%)} & \multirow{2}{*}{ $mAP$(\%) } \\
		\cline { 3 - 9 } & & Anger & Disgust & Fear & Happy & Neutral & Sad & Surprise &  \\	
	\hline SSD~\cite{Wei} & 2015 & 62.77 & 47.24 & 44.74 & $\mathbf{91.20}$ & $\mathbf{55.50}$ & 66.48 & $\underline{46.59}$ & 59.22 \\
	\cline { 1 - 10} RetinaNet~\cite{Lin2017} & 2017 & $\underline{68.91}$ & 58.59 & 55.23 & 81.87 & 43.10 & 64.16 & 24.86 & 56.67   \\
	\cline { 1 - 10} YOLOv3~\cite{Redmon2018YOLOv3AI} & 2018 & 19.52 & 0.00 & 5.88 & 50.28 & 37.45 & 21.11 & 0.00 & 19.18   \\
	\cline { 1 - 10} CenterNet~\cite{Xingyi} & 2019 & 39.54 & 0.00 & 25.12 & 68.57 & 22.14 & 43.95 & 0.00 & 28.48   \\
	\cline { 1 - 10} EfficientNet~\cite{Mingxing} & 2019 & 15.40 & 1.18 & 17.81 & 29.68 & 17.02 & 29.26 & 0.72 & 15.87    \\
	\cline { 1 - 10} YOLOv4~\cite{Bochkovskiy2020YOLOv4OS} & 2020 & 29.58 & 0.00 & 0.00 & 21.12 & 13.78 & 21.67 & 0.00 & 12.31    \\
	\cline { 1 - 10} YOLOv5 & 2020 & 23.56 & 0.00 & 0.00 & 23.64 & 25.52 & 22.71 & 0.00 & 13.63   \\
	\cline { 1 - 10} YOLOvX~\cite{Ge2021YOLOXEY} & 2021 & 67.01 & $\mathbf{73.86}$ & $\mathbf{70.48}$ & 90.81 & 36.15 & $\underline{70.26}$ & 39.55 & $\underline{64.02}$  \\
	\cline { 1 - 10} YOLOv7~\cite{Wang2022YOLOv7TB} & 2022 & 57.47 & $\underline{64.64}$ & 52.55 & 74.34 & 32.44 & 48.44 & 32.26 & 52.02   \\
	\cline { 1 - 10} YOLOv8 & 2023 & 56.24 & 45.24 & 53.76 & 87.50 & 33.48 & 44.69 & 42.68 & 51.94   \\
	\cline { 1 - 10} FER-YOLO-Mamba & 2024 & $\mathbf{74.07}$ & 64.49 & $\underline{58.87}$ & $\underline{90.94}$ & $\underline{48.01}$ & $\mathbf{71.83}$ & $\mathbf{58.52}$ & $\mathbf{66.67}$  \\
	\hline
\end{tabular}
\end{table*}

Tables \ref{table_RAF_result} and \ref{table_sfew_result} provide a comparative analysis of our proposed FER-YOLO-Mamba network model against current state-of-the-art approaches, using the RAF-DB and SFEW datasets as benchmarks. In these tables, the optimally performing results are highlighted in bold, and the next superior ones are marked with an underline for clarity.

Drawing insights from the data analysis presented in Tables \ref{table_RAF_result} and \ref{table_sfew_result}, the FER-YOLO-Mamba model has attained the highest $mAP$ scores across the two benchmark datasets, RAF-DB and SFEW, recording 80.31\% and 66.67\% respectively. This performance notably surpasses the prevailing state-of-the-art YOLOvX model, leading by 1.91\% on RAF-DB and 2.65\% on SFEW.

In particular, the FER-YOLO-Mamba model demonstrated outstanding performance in handling the ``Sad'' and ``Surprise'' emotion classes. On the RAF-DB dataset, the model achieved an $AP$ score of 84.22\% for the ``Sad'' class and 91.44\% for the ``Surprise'' class. When transitioning to the SFEW dataset, the FER-YOLO-Mamba model maintained high recognition accuracy with $AP$ scores of 71.78\% and 58.52\% for the ``Sad'' and ``Surprise'' classes respectively.

Notably, on the RAF-DB dataset, the ``Happy'' emotion class achieved an outstanding $AP$ score of 97.43\%, the highest amongst all classes. While on the SFEW dataset, the FER-YOLO-Mamba model's $AP$ for the ``Happy'' class, though marginally 0.26\% lower than that of the SSD model, remained at a competitively high level compared to other methodologies. This is mainly attributed to the richer and more diverse samples of the ``Happy'' emotion class in the dataset, providing the FER-YOLO-Mamba model with a broader learning space to better understand and differentiate the features and patterns of this emotion.

However, in a comprehensive evaluation of the performance of the FER-YOLO-Mamba model, several aspects emerged as avenues for enhancement. On the RAF-DB dataset, the model's performance was comparatively weaker in identifying the ``Fear'' emotion, as evidenced by a lower $AP$ score. Similarly, when tested on the SFEW dataset, the model's handling of the ``Neutral'' emotion class proved to be a challenge, yielding suboptimal $AP$ scores. These observations point to specific emotional classes that may require further tuning or specialized attention in future iterations of the model.

\begin{figure}[!t]
	\centering
	\includegraphics[width=3.4in]{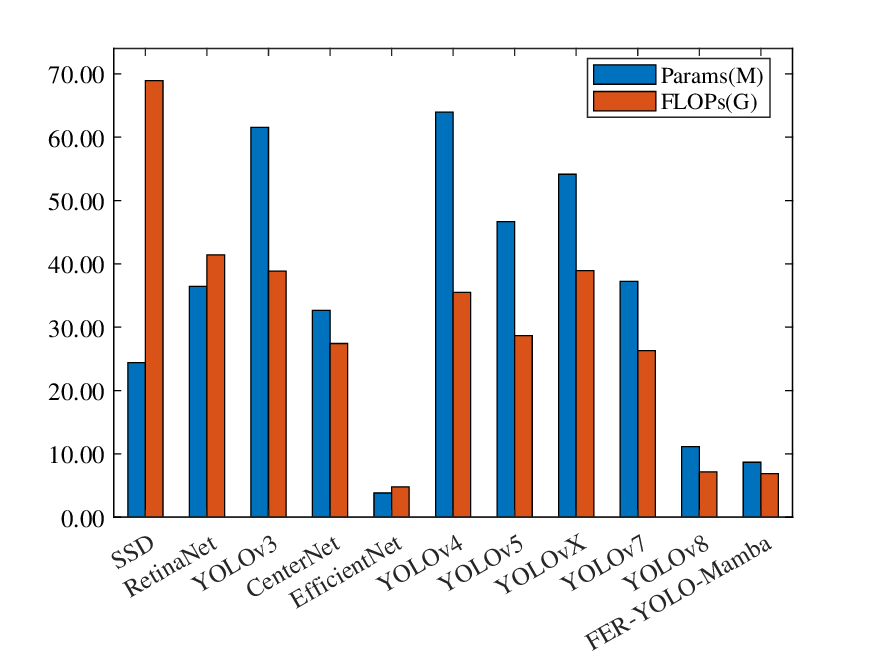}
	\caption{Comparison of different network models in terms of Params~(M) and FLOPs~(G).}
	\label{fig_pf}
\end{figure}

Further, we present a comparative analysis of the Params and FLOPs across various network models in \figref{fig_pf}. Notably, EfficientNet stands out with its extremely low number of parameters and computational cost, with Params of 3.83M and FLOPs of 4.78G, highlighting its profound efficiency in resource utilization. Following closely is our proposed FER-YOLO-Mamba model, with Params of 8.68M and FLOPs of 6.89G, slightly higher compared to EfficientNet but still maintained at a relatively low level.

Nonetheless, in assessing the models comprehensively, it is imperative to reconcile the counts of parameters and computational costs with their performance. Despite EfficientNet having lower parameter size and computational requirements than FER-YOLO-Mamba, its performance significantly lags on the two datasets in terms of $mAP$, with differences of 8.86\% on RAF-DB and 50.80\% on SFEW, which indicates that in pursuit of higher resource efficiency at the expense of a certain degree of model complexity, EfficientNet does compromise on overall classification accuracy.

\subsection{Ablation Experiments}
\begin{table*}[!t]
	\centering
	\caption{Performance of different components within FER-YOLO-Mamba Network.}
	\label{Ablation_result}
	\begin{tabular}{|c|c|c|c|c|c|c|c|c|} 
		\hline
		FRM & OSS & OSSM & SS2D\cite{liu2024vmamba} & {\it Avg\_F1} & {\it Avg\_Recall}(\%) & {\it Avg\_Precision}(\%)    & $mAP$(\%) & Dataset  \\ 
		\hline
		$\sqrt{ }$ & & & & 0.72  & 69.04  & 77.65  & 76.24  & \multirow{5}{*}{RAF-DB}  \\ 
		\cline{1-8}
		& $\sqrt{ }$ &  & $\sqrt{ }$& 0.71  & 66.55  & \textbf{80.21} & 78.58  &  \\ 
		\cline{1-8}
		$\sqrt{ }$ & $\sqrt{ }$ &  & $\sqrt{ }$& 0.72  & 68.81  & 79.51  & 78.67  &  \\ 
		\cline{1-8}
		& $\sqrt{ }$ & $\sqrt{ }$&  & 0.74  & 70.90  & 78.83  & 79.54  &  \\ 
		\cline{1-8}
		$\sqrt{ }$ & $\sqrt{ }$ & $\sqrt{ }$&  & \textbf{0.75} & \textbf{72.95} & 78.29  & \textbf{80.31} &  \\ 
		\hline
		$\sqrt{ }$ & & & & 0.59  & 59.54  & 61.45  & 64.67  & \multirow{5}{*}{SFEW}    \\ 
		\cline{1-8}
		& $\sqrt{ }$ &  & $\sqrt{ }$& \textbf{0.63} & 65.94  & 61.48  & 59.05  &  \\ 
		\cline{1-8}
		$\sqrt{ }$ & $\sqrt{ }$ &  & $\sqrt{ }$& 0.61  & 67.19  & 58.09  & 57.82  &  \\ 
		\cline{1-8}
		& $\sqrt{ }$ & $\sqrt{ }$&  & \textbf{0.63} & \textbf{68.32} & 60.12  & 58.23  &  \\ 
		\cline{1-8}
		$\sqrt{ }$ & $\sqrt{ }$ & $\sqrt{ }$&  & 0.60   & 59.60 & \textbf{63.66}  & \textbf{66.67} &  \\
		\hline
	\end{tabular}
\end{table*}
Table \ref{Ablation_result} delineates the individual contributions of various components integrated within the FER-YOLO-Mamba network, assessed across the RAF-DB and SFEW datasets. These components encompass FRM, OSS, its variant OSSM, and the SS2D.

In Table \ref{Ablation_result}, $Avg\_Precision$ represents the average precision across all emotion classes. $Avg\_Precision$ provides a comprehensive evaluation reflecting the model's prediction accuracy across all classes. Conversely, $Avg\_Recall$ denotes the average recall across classes, measuring the model's success in identifying true positive cases amidst all actual positives. A high recall implies a robust capacity to detect positives, underscoring the model's sensitivity in recognizing various emotions. Lastly, $Avg\_F1$ represents the average $F1\text{ \it score}$ across classes, which integrates precision and recall into a single measure. The $F1\text{ \it score}$ acts as a balanced indicator of a model's performance, rewarding models that excel in both avoiding false positives and capturing true positives effectively. A high $F1\text{ \it score}$ indicates a good balance between precision and recall.

Table \ref{Ablation_result} illustrates that, in the context of the RAF-DB dataset, enabling FRM, OSS, and OSSM yields a relatively high $mAP$ score, highlighting the ensemble's robust performance across this dataset. The elevated $Avg\_F1$ score suggests that the model has achieved a quite good balance between $Precision$ and $Recall$. Furthermore, the high $Avg\_Recall$ underscores the network's efficacy in distinguishing the majority of positive cases, affirming its strength in detection.

Conversely, on the SFEW dataset, while the $mAP$ remains relatively high, it suggests a consistent overall performance. A lower $Avg\_F1$ implies that the model's balance between $Precision$ and $Recall$ is not ideal. A higher $Avg\_Precision$ may indicate that the network faces some challenges in comprehensively retrieving all positive samples. The model tends to make predictions only when it is highly confident that a sample is positive, thus ensuring higher precision. However, this conservative and cautious strategy may lead the model to miss instances that are actually positive but not significant enough in some cases, further resulting in a decrease in $Avg\_Recall$.

\subsection{Experimental results across different classes} 
Table \ref{table_Evaluation} shows the performance of YOLOvX and FER-YOLO-Mamba across classes in the RAF-DB and SFEW datasets, where the optimally performing results are highlighted in bold.

From the data in Table \ref{table_Evaluation}, it is evident that YOLOvX and FER-YOLO-Mamba exhibit different performance levels across various emotion classes in the RAF-DB dataset. Overall, FER-YOLO-Mamba outperforms YOLOvX in terms of $F1\text{ \it score}$, $Recall$, $Precision$ and the mean of these metrics for the majority of classes, thereby demonstrating superior capability in executing FER tasks. More specifically, in the ``Anger'' and ``Surprise'' classes, while YOLOvX attains $Recall$ rates of 62.34\% and 81.29\% alongside $Precision$ rates of 81.36\% and 85.80\%, respectively, FER-YOLO-Mamba demonstrates enhanced $Recall$ rates of 71.43\% and 85.38\% and $Precision$ rates of 75.34\% and 83.43\%. Although YOLOvX slightly outperforms in $Precision$, FER-YOLO-Mamba performs better in terms of $Recall$ rate, suggesting that FER-YOLO-Mamba is less prone to overlooking genuine ``Anger'' and ``Surprise'' samples, albeit potentially at the cost of increased false positives.

Except for the ``Sad'' class, FER-YOLO-Mamba has a higher $F1\text{ \it score}$, denoting an enhanced performance in recognizing emotions other than ``Sad''. It is noteworthy that in both the YOLOvX and FER-YOLO-Mamba architectures, the ``Happy'' emotion class achieves exceptional recognition accuracy across both datasets. This may be attributed to the distinctive features of happy expressions, facilitating a heightened level of accuracy and reliability in their identification.

However, in the RAF-DB dataset, the $F1\text{ \it score}$ for the ``Disgust'' and ``Fear'' classes is relatively low, indicating that these expressions are more challenging for the models. This could be due to the subtle nature of these expressions, significant individual variations, or the potential for confusion with other expressions. Similarly, in the SFEW dataset, the $F1\text{ \it score}$ for the ``Surprise'' and ``Neutral'' classes is also low, suggesting that the models face significant difficulties in recognizing these expressions. This may be because the features of these expressions are similar to other expressions, posing challenges for the models in differentiation.
\begin{table*}
	\centering
	\caption{Performance comparison of YOLOvX and FER-YOLO-Mamba across classes in the RAF-DB and SFEW datasets.}  
	\label{table_Evaluation} 
	\begin{tabular}{|c|c|c|c|c|c|c|c|c|c|} 
		\hline
		\multirow{2}{*}{Class} & \multicolumn{4}{c|}{YOLOvX}                                                                      & \multicolumn{5}{c|}{FER-YOLO-Mamba}                                                                              \\ 
		\cline{2-10}
		& F1 score       & Recall(\%)                               & Precision(\% )& $mAP$(\%)               & F1 score       & Recall(\%)       & Precision(\%)    & $mAP$(\%)                         & Dataset                  \\ 
		\hline
		Anger                  & 0.71           & 62.34                                   & 81.36        & \multirow{8}{*}{78.40} & 0.73           & 71.43           & 75.34           & \multirow{8}{*}{\textbf{80.31} } & \multirow{8}{*}{RAF-DB}  \\ 
		\cline{1-4}\cline{6-8}
		Disgust                & 0.59           & 52.38                                   & 64.29        &                        & 0.61           & 53.33           & 70.00           &                                  &                          \\ 
		\cline{1-4}\cline{6-8}
		Fear                   & 0.57           & 51.43                                   & 64.29        &                        & 0.61           & 48.57           & 80.95           &                                  &                          \\ 
		\cline{1-4}\cline{6-8}
		Happy                  & 0.92           & 93.39                                   & 90.28        &                        & 0.91           & 92.53           & 90.45           &                                  &                          \\ 
		\cline{1-4}\cline{6-8}
		Neutral                & 0.74           & 82.51                                   & 67.39        &                        & 0.75           & 84.41           & 67.89           &                                  &                          \\ 
		\cline{1-4}\cline{6-8}
		Sad                    & 0.79           & 75.48                                   & 83.51        &                        & 0.77           & 75.00           & 80.00           &                                  &                          \\ 
		\cline{1-4}\cline{6-8}
		Surprise               & 0.83           & 81.29                                   & 85.80        &                        & 0.84           & 85.38           & 83.43           &                                  &                          \\ 
		\cline{1-4}\cline{6-8}
		Average                & 0.74           & \textasciitilde{}71.26\textasciitilde{} & 76.70        &                        & \textbf{0.75}  & \textbf{72.95}  & \textbf{78.29}  &                                  &                          \\ 
		\hline
		Class                  & F1 score       & Recall(\%)                               & Precision(\%) & $mAP$(\%)               & F1 score       & Recall(\%)       & Precision(\%)    & $mAP$(\%)                         & Dataset                  \\ 
		\hline
		Anger                  & 0.62           & 69.57                                   & 55.17        & \multirow{8}{*}{64.02} & 0.65           & 78.26           & 56.25           & \multirow{8}{*}{\textbf{66.7} }  & \multirow{8}{*}{SFEW}    \\ 
		\cline{1-4}\cline{6-8}
		Disgust                & 0.67           & 85.71                                   & 54.55        &                        & 0.62           & 57.14           & 66.67           &                                  &                          \\ 
		\cline{1-4}\cline{6-8}
		Fear                   & 0.64           & 52.94                                   & 81.82        &                        & 0.48           & 41.18           & 58.33           &                                  &                          \\ 
		\cline{1-4}\cline{6-8}
		Happy                  & 0..85          & 88.00                                   & 81.48        &                        & 0.88           & 92.00           & 85.19           &                                  &                          \\ 
		\cline{1-4}\cline{6-8}
		Neutral                & 0.43           & 38.10                                   & 50.00        &                        & 0.42           & 33.33           & 58.33           &                                  &                          \\ 
		\cline{1-4}\cline{6-8}
		Sad                    & 0.62           & 66.67                                   & 57.14        &                        & 0.71           & 70.83           & 70.83           &                                  &                          \\ 
		\cline{1-4}\cline{6-8}
		Surprise               & 0.48           & 66.67                                   & 37.50        &                        & 0.47           & 44.44           & 50.00           &                                  &                          \\ 
		\cline{1-4}\cline{6-8}
		Average                & \textbf{0.62}  & \textbf{66.81}                          & 59.67        &                        & 0.60           & 59.60           & \textbf{63.66}  &                                  &                          \\
		\hline
	\end{tabular}
\end{table*}

\subsection{Visualization of the detection results}
\begin{figure*}[!t]
	\centering
	\scriptsize
	\renewcommand{\tabcolsep}{0pt}
	\begin{tabular}{cccc}
	\includegraphics[width=1in,height=1in]{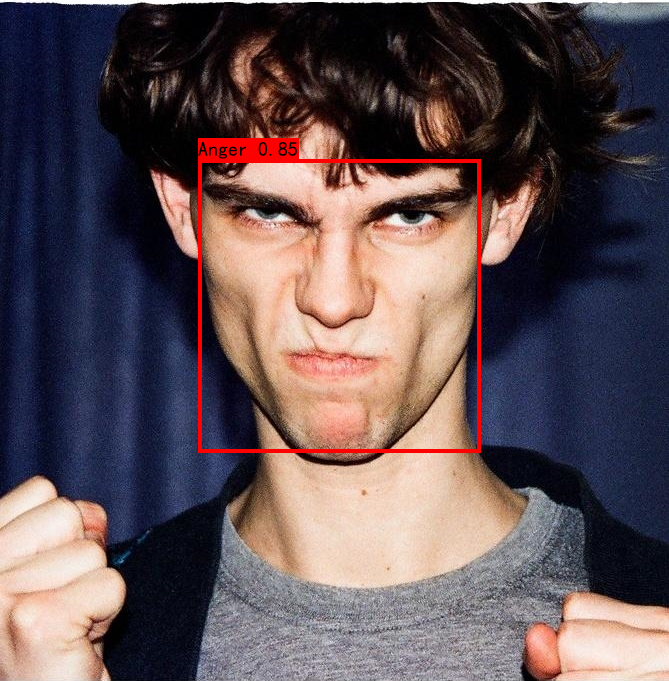} &
	\includegraphics[width=1in,height=1in]{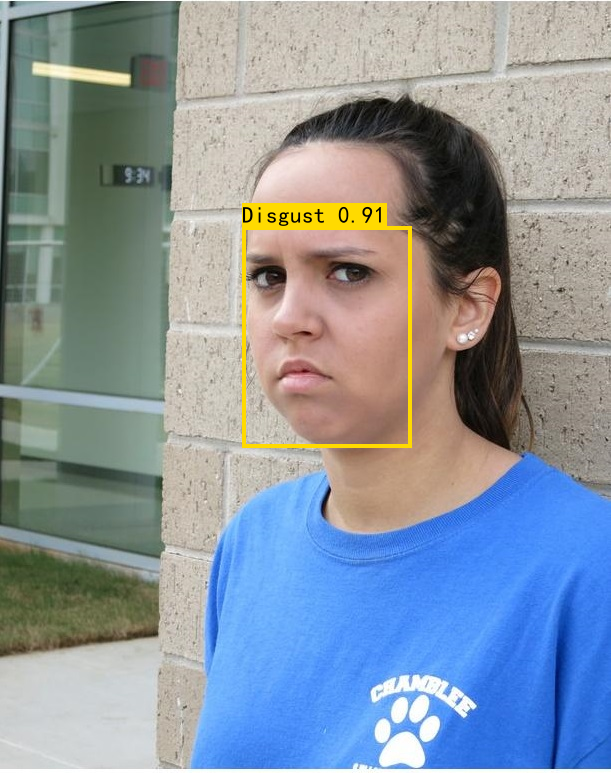} &
	\includegraphics[width=1in,height=1in]{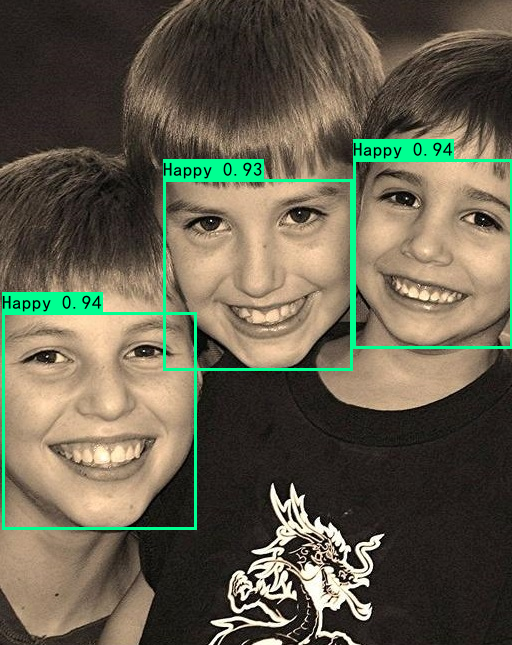} &
	\includegraphics[width=1in,height=1in]{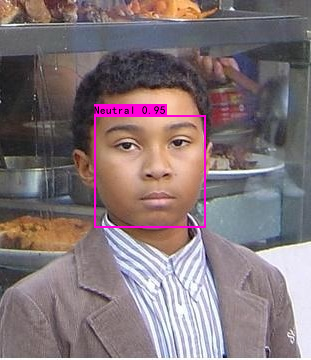} \\
	\includegraphics[width=1in,height=1in]{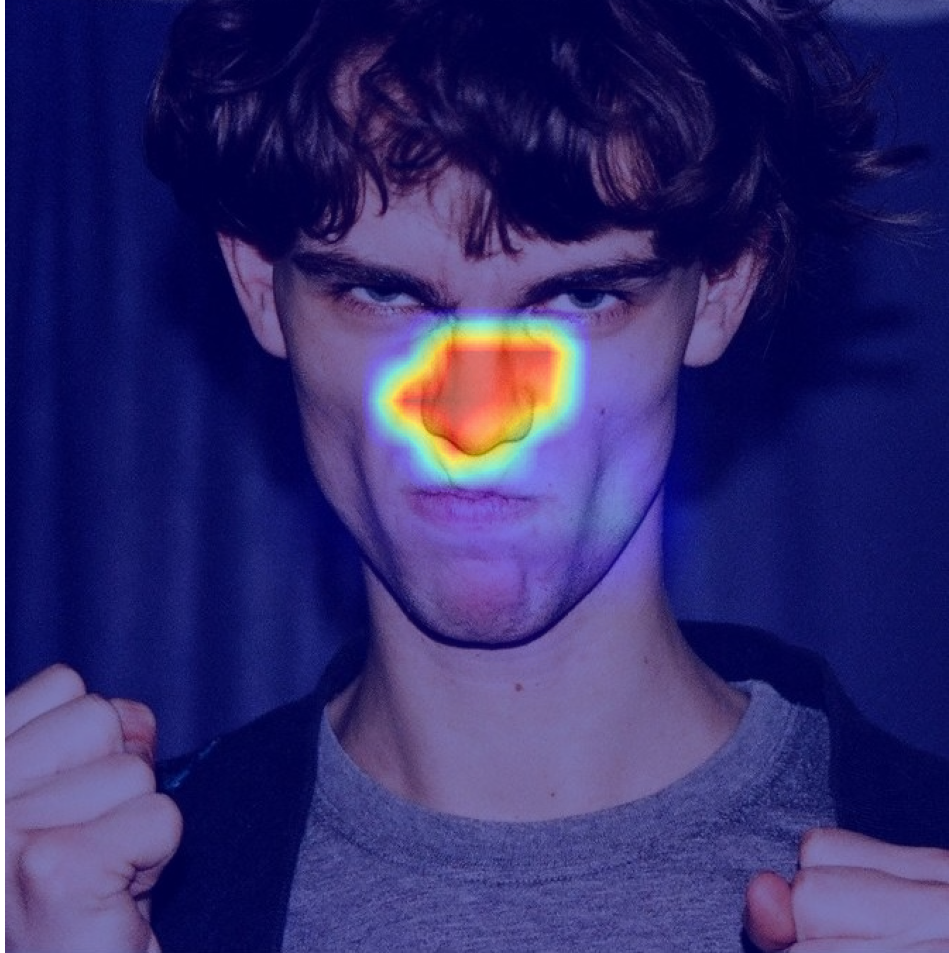} &
	\includegraphics[width=1in,height=1in]{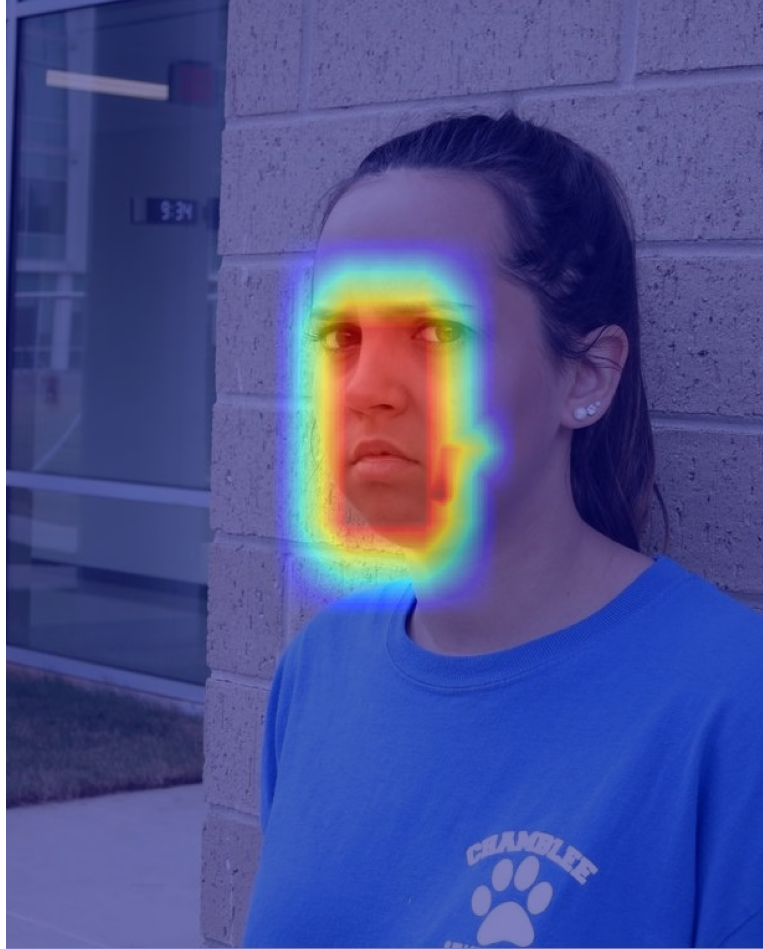} &
	\includegraphics[width=1in,height=1in]{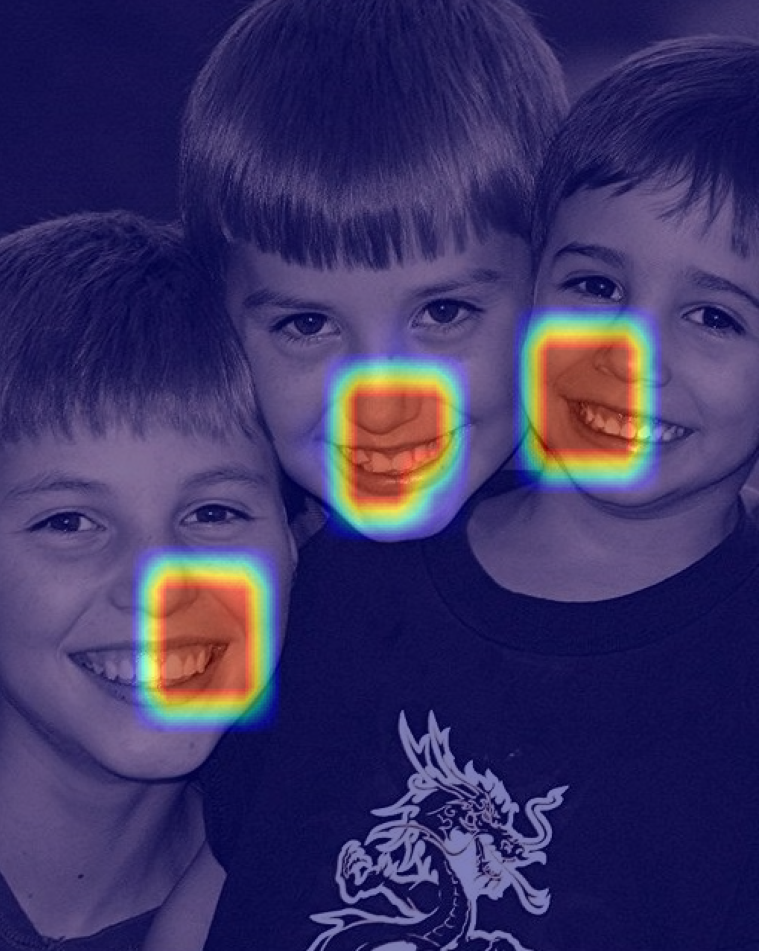} &
	\includegraphics[width=1in,height=1in]{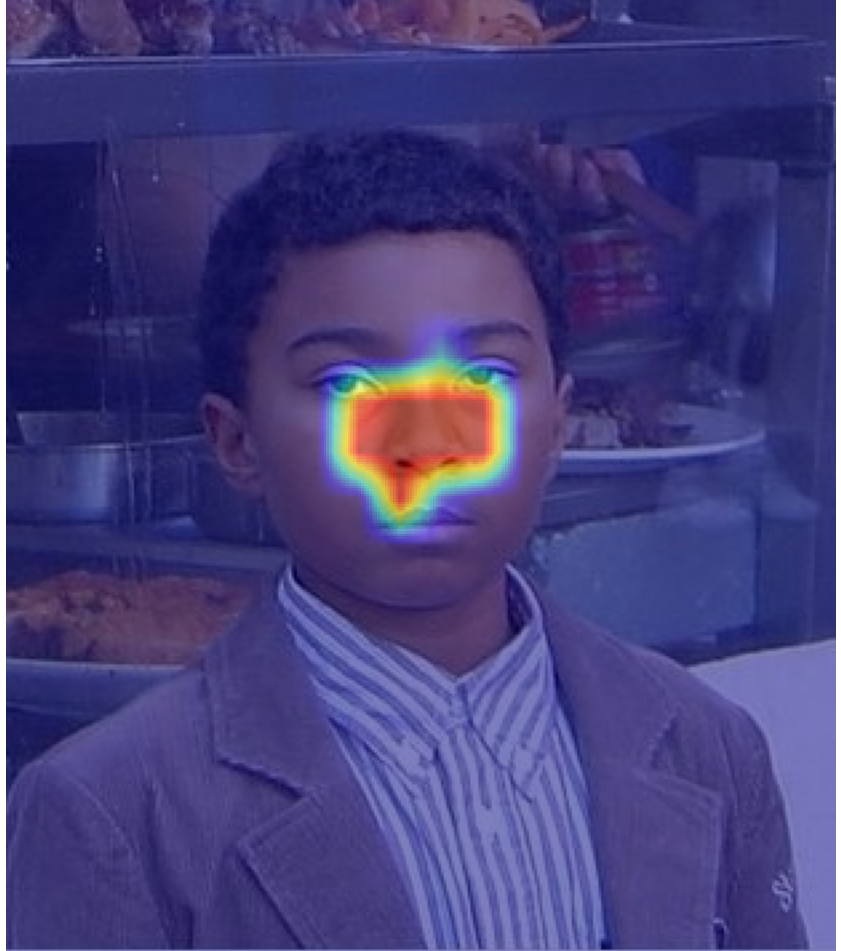} \\
	(a) Anger & (b) Disgust & (c) Happy & (d) Neutral
	\end{tabular}
	\begin{tabular}{ccc}
		\includegraphics[width=1in,height=1in]{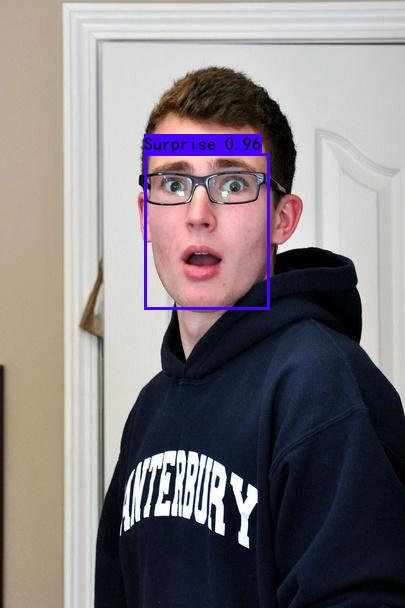} &
		\includegraphics[width=1in,height=1in]{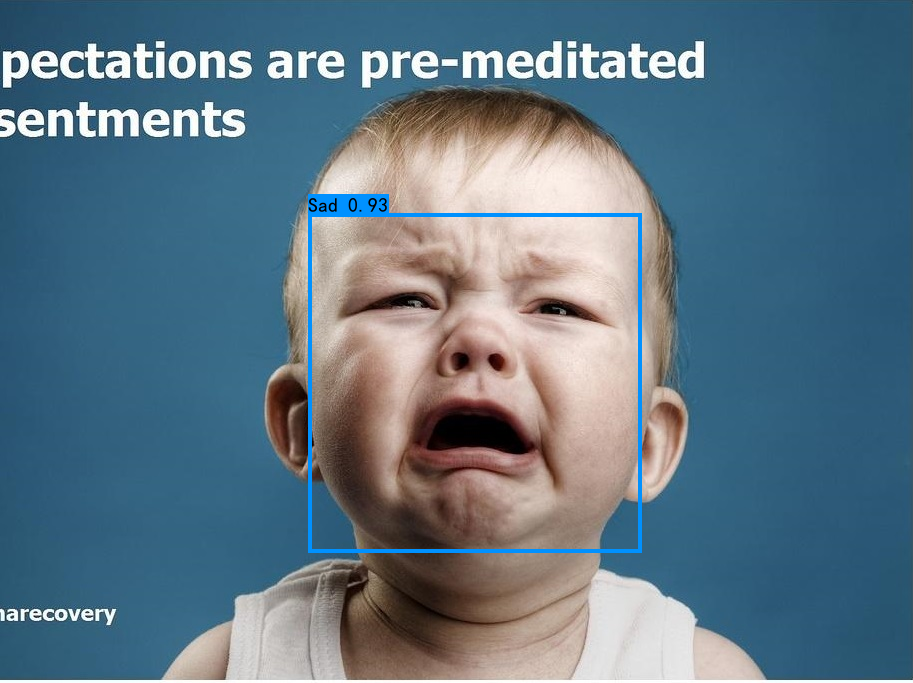} &
		\includegraphics[width=1in,height=1in]{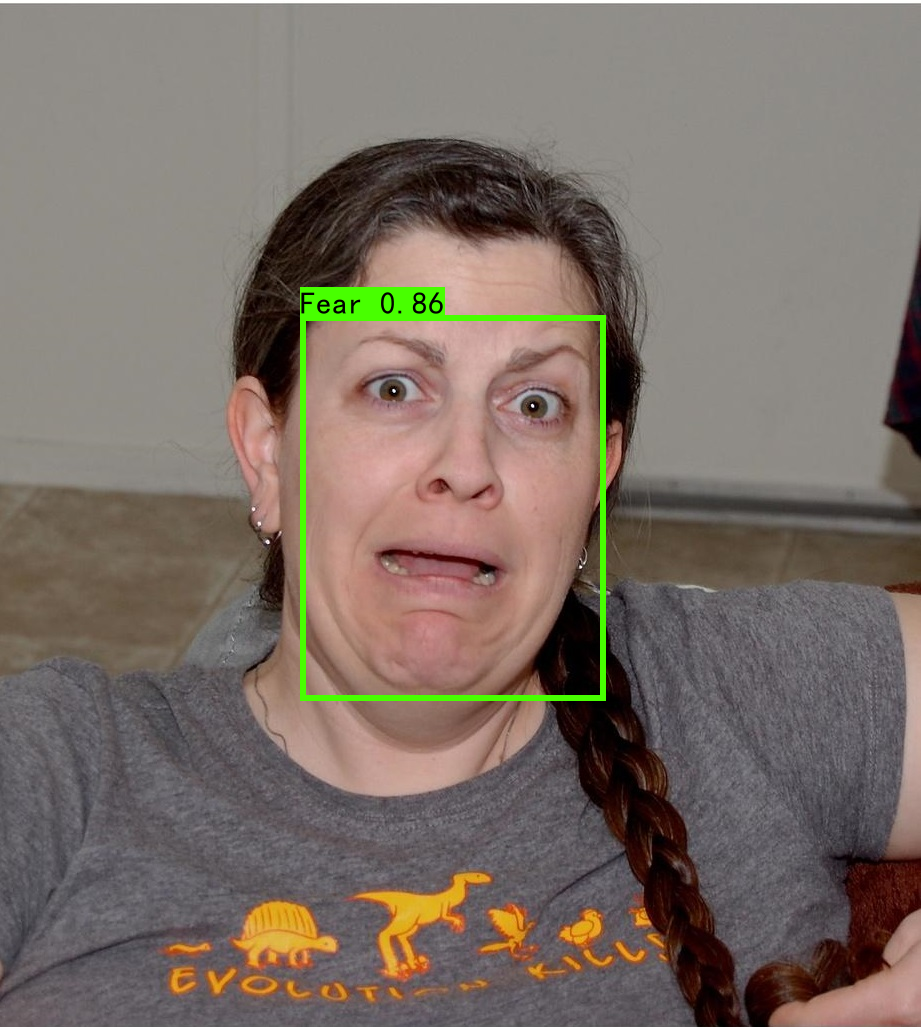} \\
		\includegraphics[width=1in,height=1in]{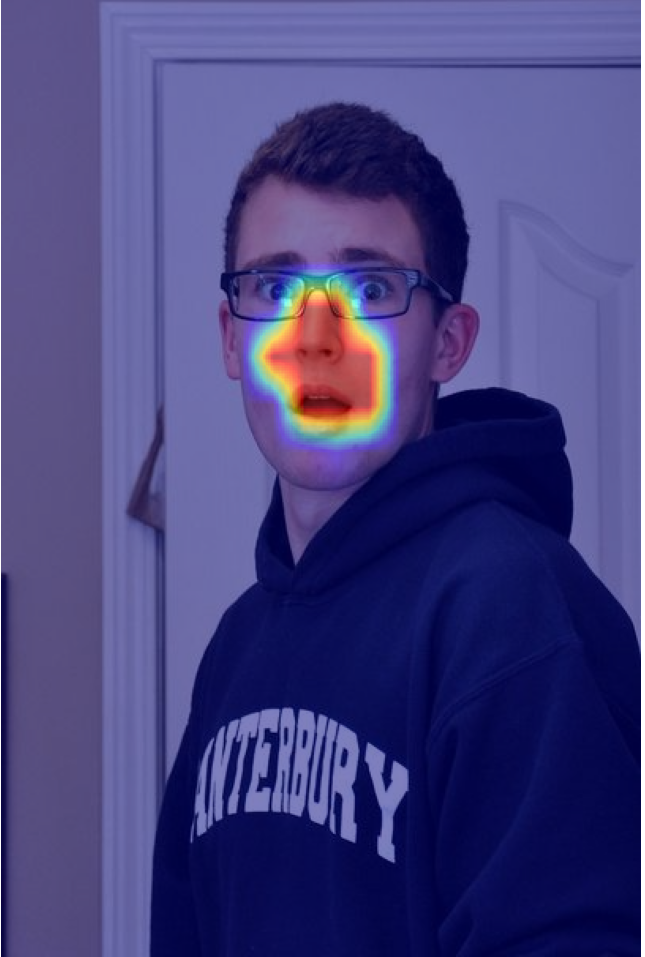} &
		\includegraphics[width=1in,height=1in]{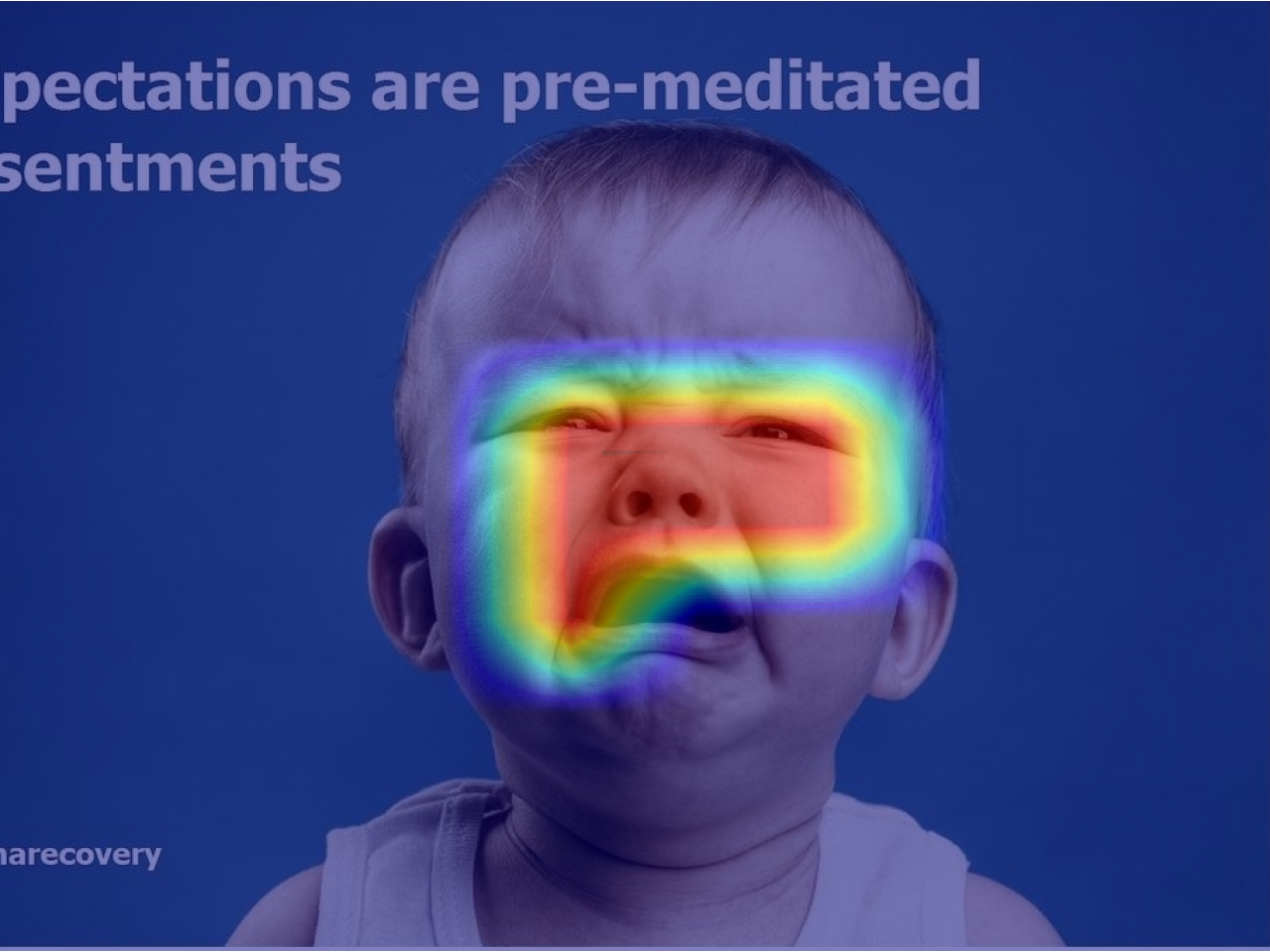} &
		\includegraphics[width=1in,height=1in]{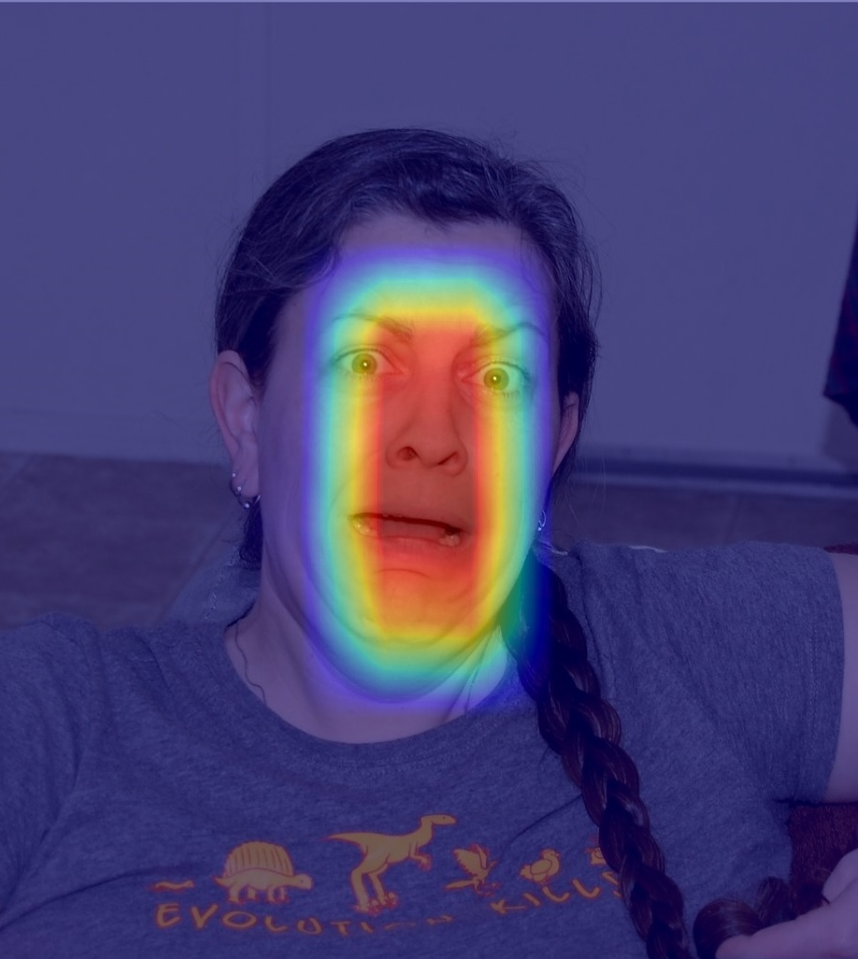} \\
		(e) Surprise  & (f) Sad & (g) Fear
	\end{tabular}
	\caption{Test sample detection results and corresponding heatmaps on RAF-DB.}
	\label{fig_detection_results_on_RAF_DB}
\end{figure*}
\begin{figure*}[!t]
	\centering
	\scriptsize
	\renewcommand{\tabcolsep}{0pt}
	\begin{tabular}{cccc}
		\includegraphics[width=1in,height=1in]{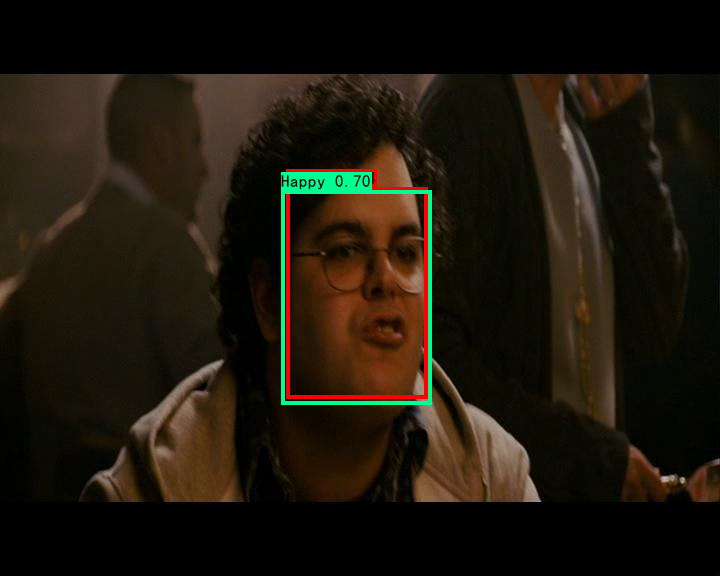} &
		\includegraphics[width=1in,height=1in]{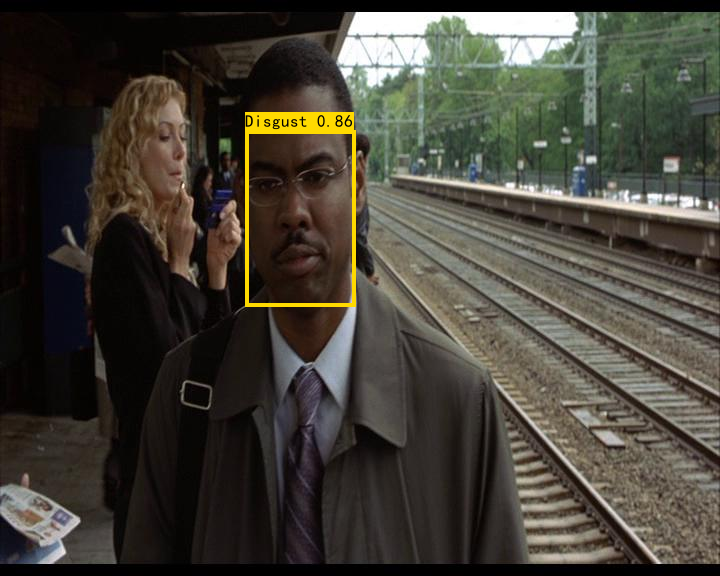} &
		\includegraphics[width=1in,height=1in]{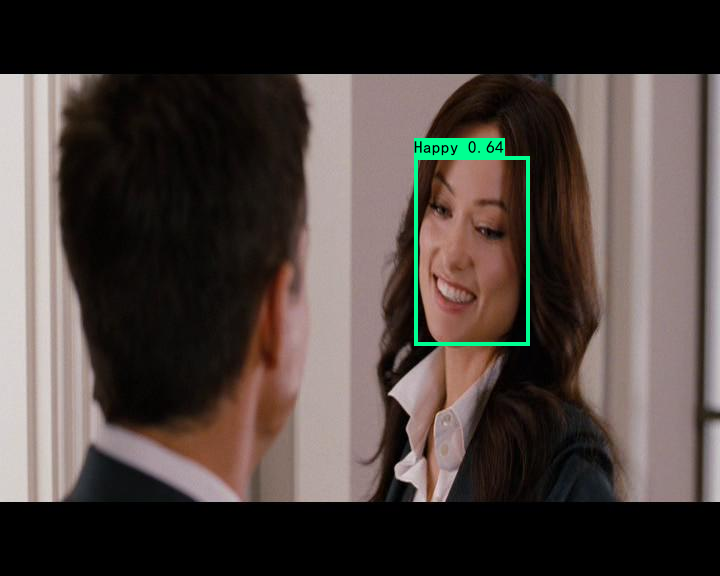} &
		\includegraphics[width=1in,height=1in]{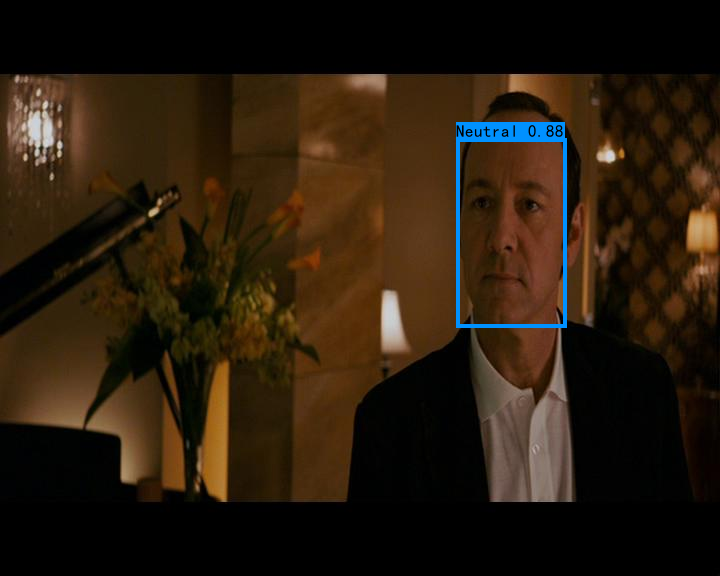} \\
		\includegraphics[width=1in,height=1in]{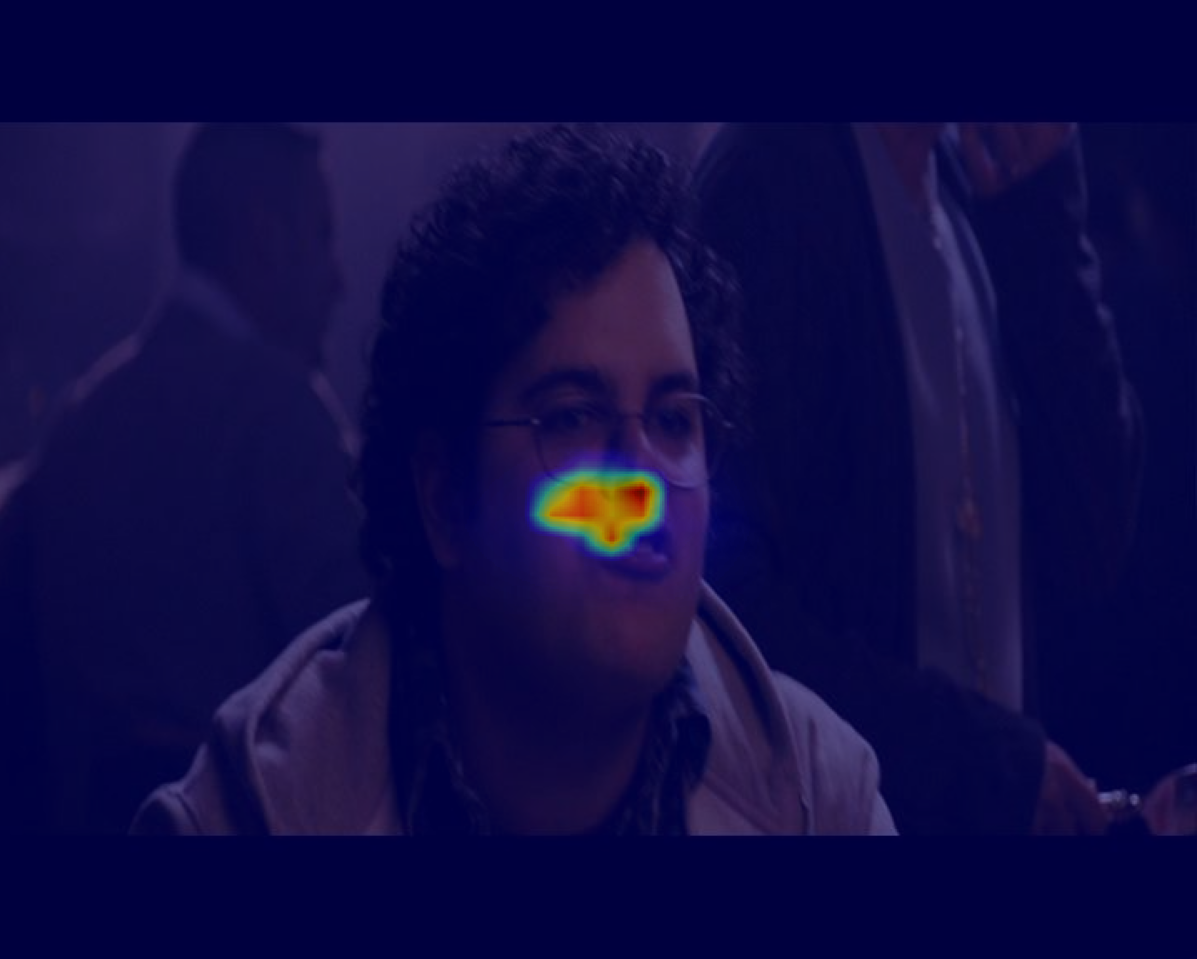} &
		\includegraphics[width=1in,height=1in]{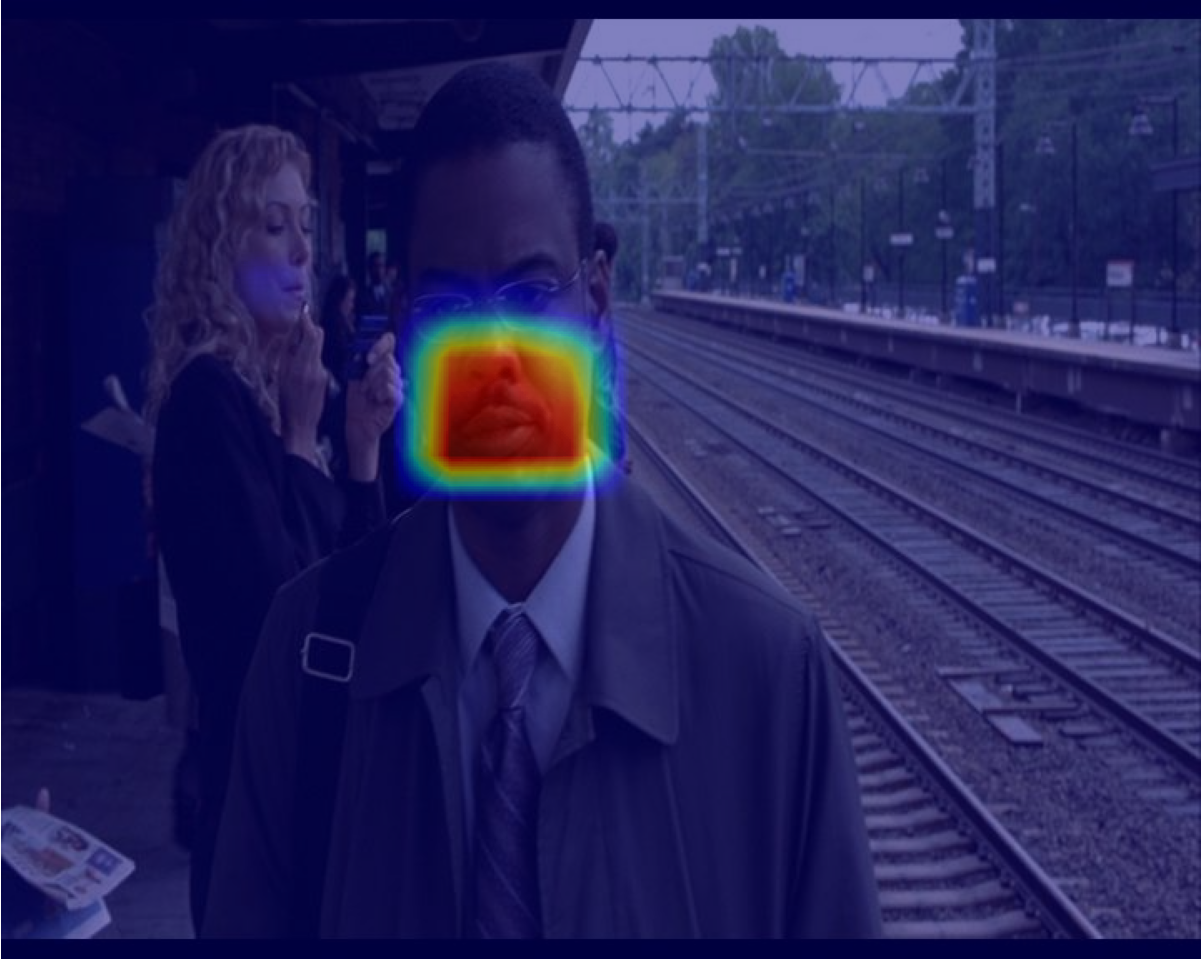} &
		\includegraphics[width=1in,height=1in]{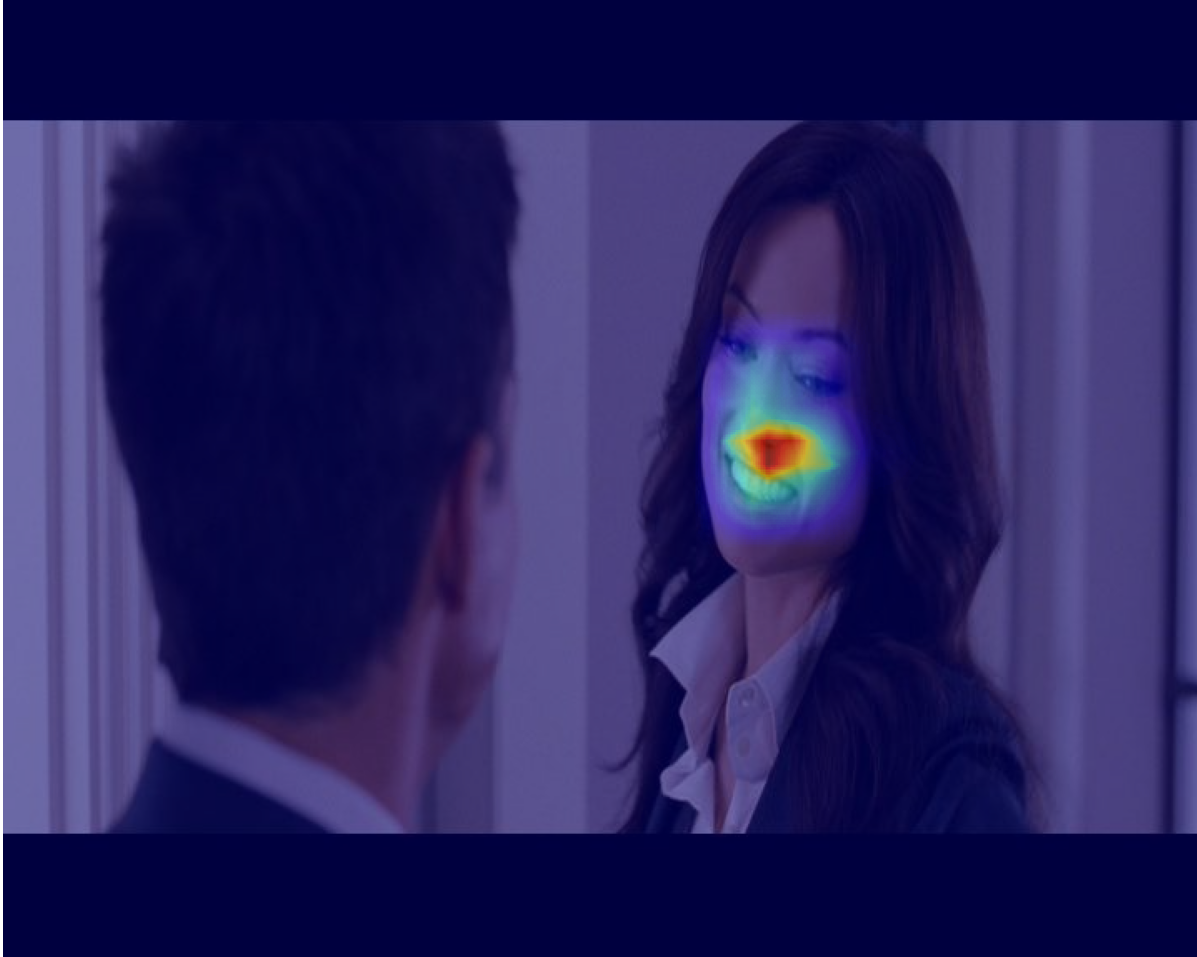} &
		\includegraphics[width=1in,height=1in]{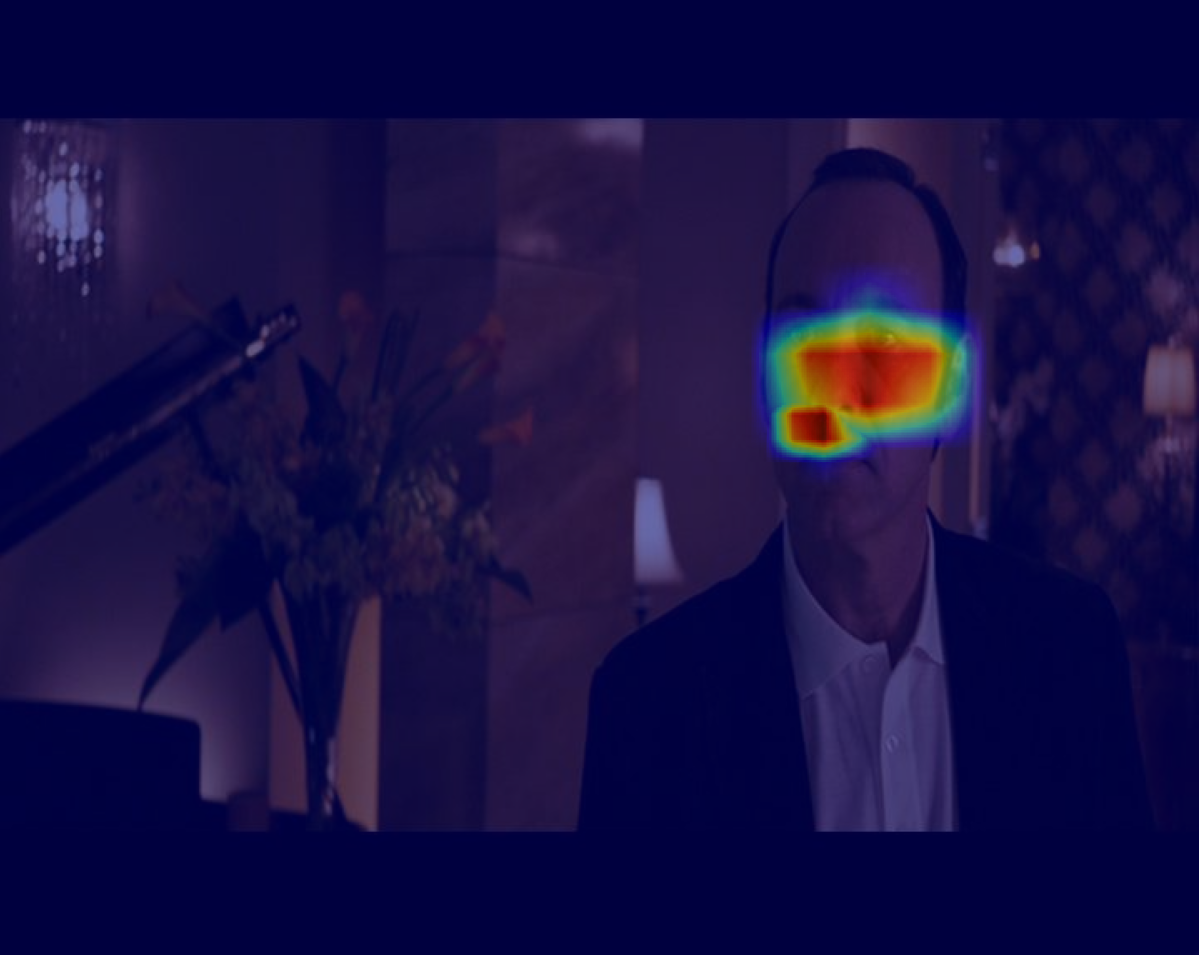} \\
		(a) Anger & (b) Disgust & (c) Happy & (d) Neutral
	\end{tabular}
	\begin{tabular}{ccc}
		\includegraphics[width=1in,height=1in]{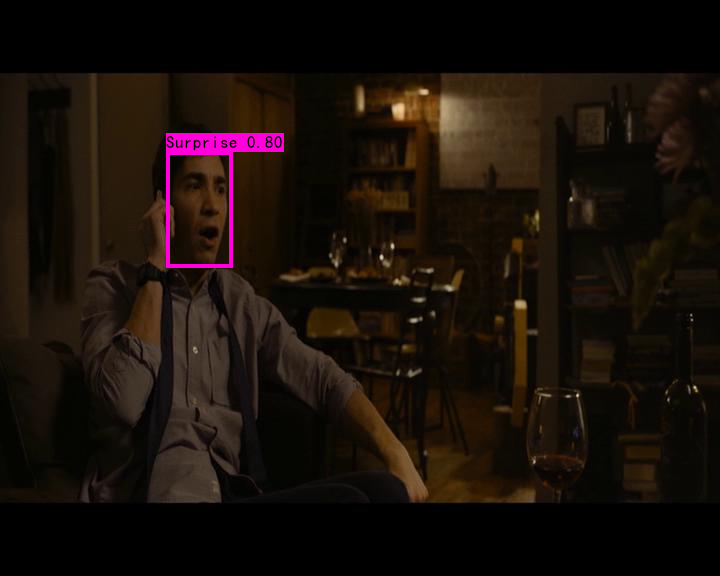} &
		\includegraphics[width=1in,height=1in]{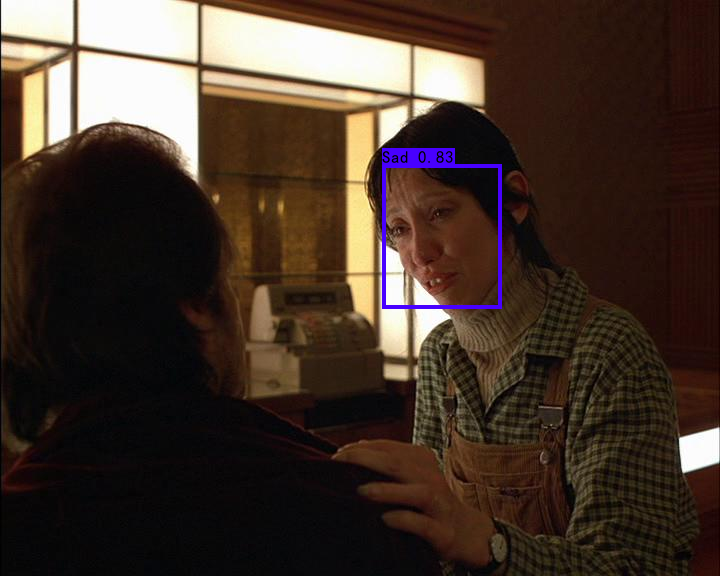} &
		\includegraphics[width=1in,height=1in]{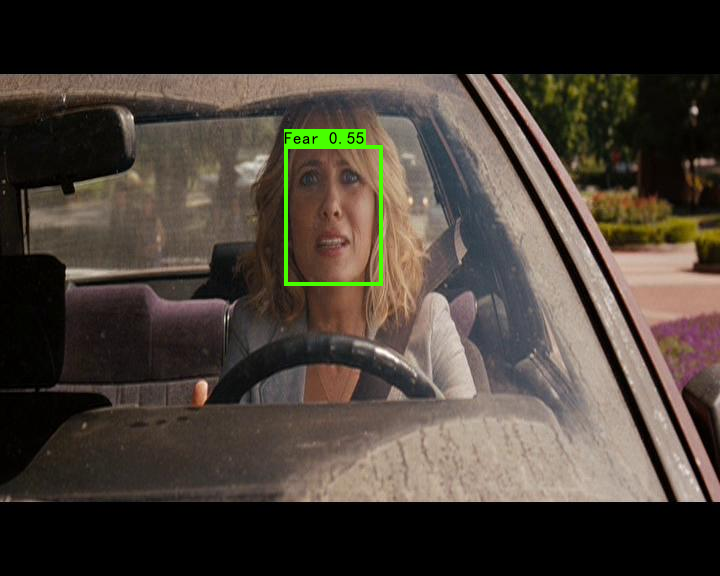} \\
		\includegraphics[width=1in,height=1in]{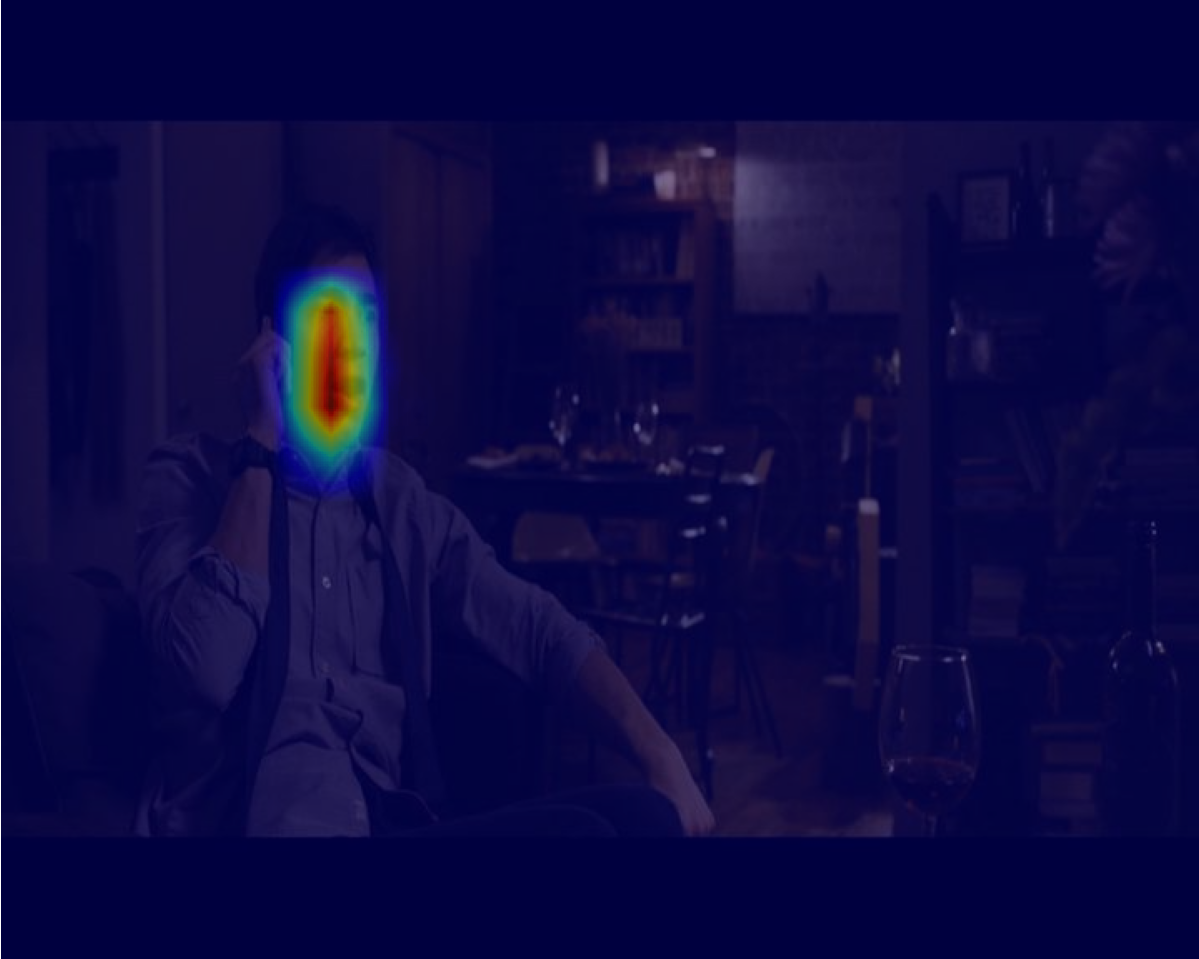} &
		\includegraphics[width=1in,height=1in]{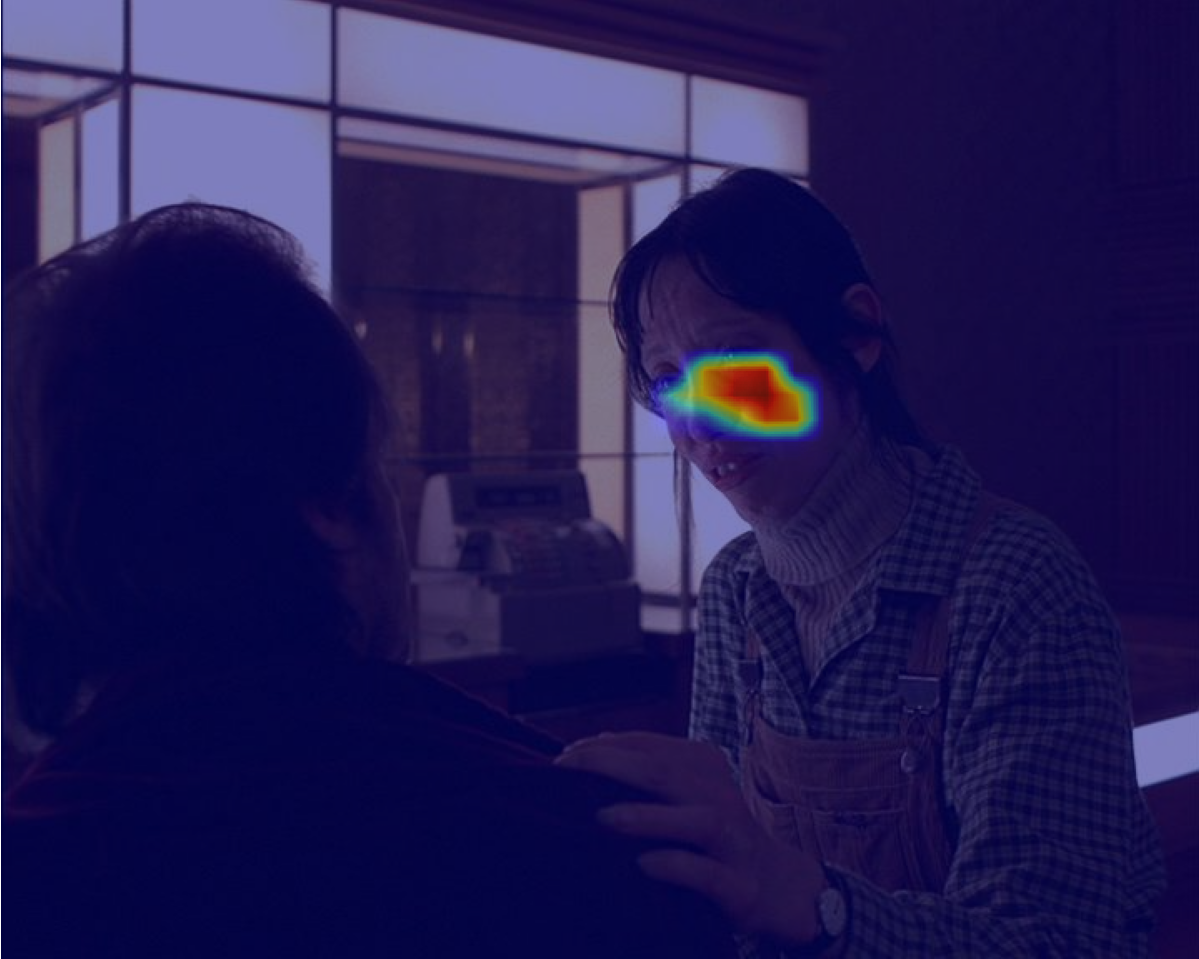} &
		\includegraphics[width=1in,height=1in]{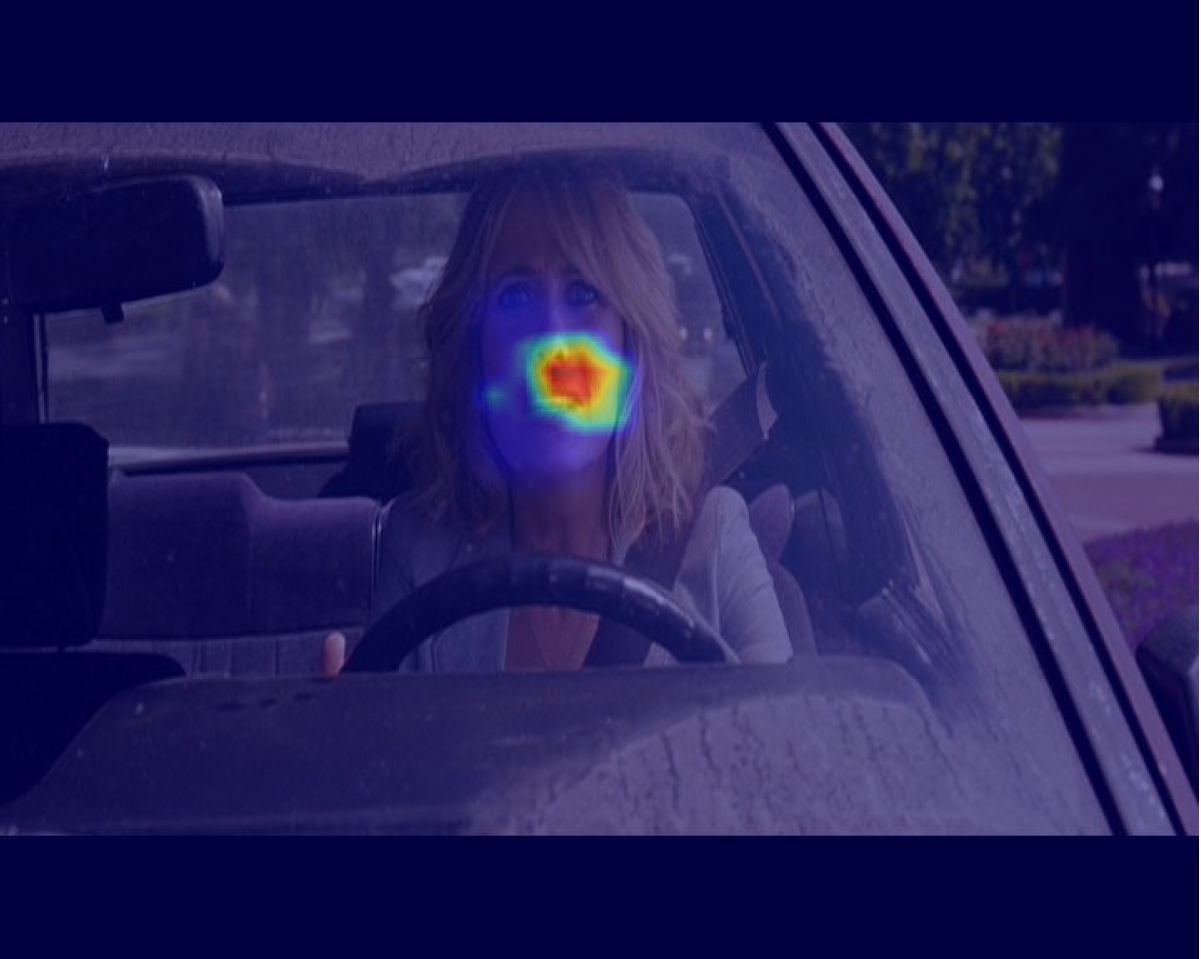} \\
		(e) Surprise  & (f) Sad & (g) Fear
	\end{tabular}
	\caption{Test sample detection results and corresponding heatmaps on SFEW.}
	\label{fig_detection_results_on_SFEW}
\end{figure*}
\figref{fig_detection_results_on_RAF_DB} and \figref{fig_detection_results_on_SFEW} display visualizations of the detection results and corresponding heatmaps obtained by applying the FER-YOLO-Mamba network model for facial expression detection on the RAF-DB and SFEW benchmark datasets, respectively. In each dataset, the first row of data shows the corresponding detection results. Each image displays a human face, with the facial area highlighted by colored bounding boxes. The boxes also display the predicted emotion class along with their confidence scores. For example, ``Anger 0.85'' indicates that the system identifies the facial expression as ``Anger'' with a confidence score of 0.85. Following the first-row detection results, the second row displays the corresponding heatmap representation.

The generation principle of heatmaps is to map the values in a two-dimensional data matrix to colors and fill these colors into the corresponding coordinate grid to form a visually comprehensible image. This visualization method helps reveal the distribution, aggregation, and correlation characteristics of data in two dimensions. The most intuitive effect of heat maps is to present the distribution of data through color, making the characteristics of the data clear at a glance. The color depth or hue change of the grid cells reflects the size or density of the data values at that location. Areas with darker colors represent higher values or greater density, whereas regions with lighter colors indicate lower values or lesser density.

The visualizations demonstrate the robust FER capabilities of the FER-YOLO-Mamba network even in image scenes with complex background interference. The model accurately locates the facial area and effectively extracts crucial facial features from visual noise, enabling precise identification and annotation of the individual's emotional state. These observations confirm the model's effective capture of various expressions and its robust recognition capabilities in practical application conditions.

\section{Conclusion}
\label{sec:Conclusion}

Aiming to address the complexity and overhead associated with traditional FER approaches, this paper proposed a YOLO-based solution to alleviate the burdensome preprocessing, feature extraction, and classification stages characteristic of traditional visual-based FER methods. Furthermore, a FER-YOLO-Mamba network model combining a state space model was proposed, which effectively integrated the efficient feature extraction capabilities of deep learning with the ability of the state space model to capture long-range dependencies. Experimental results on RAF-DB and SFEW datasets demonstrated the robust performance and generalization ability of the proposed FER-YOLO-Mamba model in FER tasks, effectively addressing various contexts and complexities presented by these challenging datasets.

\section*{Acknowledgments}
This work was supported in part by the National Natural Science Foundation of China under Grant 62271418, and in part by the Natural Science Foundation of Sichuan Province under Grant 2023NSFSC0030.

\bibliographystyle{IEEEtran}
\bibliography{FER_YOLO_Mamba}

\vfill

\end{document}